\documentclass{article}

\usepackage[english]{babel}

\usepackage[letterpaper,top=2cm,bottom=2cm,left=3cm,right=3cm,marginparwidth=1.75cm]{geometry}

\usepackage{amsmath}
\usepackage{amsfonts}
\usepackage{amssymb}
\usepackage{booktabs}
\usepackage{array}
\usepackage{authblk}
\usepackage{graphicx}
\usepackage[colorlinks=true, allcolors=blue]{hyperref}
\usepackage{enumitem}
\usepackage{color}
\usepackage{subcaption}
\usepackage{ifthen}
\usepackage{comment}
\usepackage{longtable}
\usepackage{pdflscape}
\usepackage{makecell}
\usepackage{ragged2e}
\usepackage{multirow}
\usepackage{siunitx}
\usepackage{tabularx}
\usepackage{xurl}

\usepackage{float}
\usepackage[section]{placeins}
\usepackage{afterpage}
\usepackage{flafter}

\renewcommand{\arraystretch}{1.12}
\usepackage[utf8]{inputenc}
\usepackage[T1]{fontenc}
\usepackage{needspace}

\newcolumntype{L}[1]{>{\RaggedRight\arraybackslash}p{#1}}
\newcolumntype{C}[1]{>{\Centering\arraybackslash}p{#1}}
\newcolumntype{R}[1]{>{\RaggedLeft\arraybackslash}p{#1}}

\title{Tabular foundation models for robust calibration of near-infrared chemical sensing data}

\author[1,2]{Robin Reiter}
\author[1,2]{Denis Cornet}
\author[3]{Fabien Michel}
\author[1,2]{Lauriane Rouan}
\author[1,2]{Gregory Beurier}

\affil[1]{CIRAD, UMR AGAP Institut, F-34398 Montpellier, France}
\affil[2]{UMR AGAP Institut, Université de Montpellier, CIRAD, INRAE, Institut Agro, Montpellier, France}
\affil[3]{LIRMM, Université de Montpellier, CNRS, Montpellier, France}

\begin{document}

\newboolean{showcomments}
\setboolean{showcomments}{true}
\ifthenelse{\boolean{showcomments}}
{ \newcommand{\mynote}[3]{
     \fbox{\bfseries\sffamily\scriptsize#1}
       {\small$\blacktriangleright$\textsf{\textcolor{#3}{{\em #2}\bf }}$\blacktriangleleft$}}}
       { \newcommand{\mynote}[2]{}}

\newcommand{\denis}[1]{\mynote{Denis}{#1}{magenta}}
\newcommand{\lauriane}[1]{\mynote{Lauriane}{#1}{blue}}
\newcommand{\fab}[1]{\mynote{Fab}{#1}{red}}
\newcommand{\robin}[1]{\mynote{Robin}{#1}{brown}}
\newcommand{\greg}[1]{\mynote{Greg}{#1}{orange}}

\maketitle 

\begin{abstract}
Near-infrared spectroscopy is increasingly used as a rapid, non-destructive chemical sensing technology for the analysis of food, pharmaceutical, biological, and environmental samples. However, the practical deployment of NIR sensors still depends on calibration models able to handle high-dimensional, collinear spectra, limited sample sizes, preprocessing dependence, spectral outliers, and extrapolation beyond the calibration domain. Here, we evaluate whether tabular foundation models can provide a new calibration strategy for NIR chemical sensing. We benchmark TabPFN on 66 NIR datasets covering 54 regression and 12 classification tasks, and compare direct inference on raw spectra with preprocessing-optimized inference against PLS/PLS-DA, Ridge, Catboost, and one-dimensional convolutional neural networks. The study uses a unified validation framework in which preprocessing and model selection are performed exclusively on calibration data before external test evaluation. In regression, preprocessing-optimized TabPFN achieves the best overall average rank and significantly outperforms PLS, CatBoost, TabPFN on raw spectra, and CNN-1D, while remaining statistically comparable to Ridge. In classification, TabPFN applied directly to raw spectra provides the best average rank, with performance close to the optimized variant. Robustness analyses show that TabPFN provides strong average predictive performance but that its advantage decreases on spectral outliers and extrapolated samples, where classical chemometric models remain competitive. These results suggest that tabular foundation models can complement established chemometric workflows for NIR chemical sensing, especially in small- to medium-sized calibration settings, while highlighting the need for spectroscopy-specific priors and uncertainty-aware deployment strategies.
\end{abstract}

\section{Introduction}

Near-infrared spectroscopy (NIRS) is widely used as a rapid and non-destructive analytical technique for the characterization of complex samples across domains such as agri-food analysis \cite{nirs_food_analysis, Pandiselvam2022NIRFood}, pharmaceutical manufacturing \cite{jamrogiewicz_application_2012}, biomedical sensing \cite{nirs_biomedical_sensing}, and environmental monitoring \cite{nirs_environmental_monitoring}. In these applications, the central objective is to convert spectral measurements into reliable predictions of chemical composition, physical properties, or class membership through appropriate calibration models. 

~

Indeed, NIRS data present several statistical challenges. Spectra are typically high-dimensional, strongly collinear, and often available only for relatively small sample sizes. In addition, the measured signal is affected by multiple physical perturbations such as scattering, path-length variability, and instrumental drift. These characteristics make model design and preprocessing critical components of spectroscopic analysis \cite{Rinnan2009Preprocessing}.

~

Historically, NIRS calibration has relied on chemometric models such as Partial Least Squares (PLS) \cite{Wold2001PLS} and its variants, which remain widely used due to their robustness, interpretability, and ability to handle collinearity through latent variable projections. More recently, machine learning approaches have been introduced to capture complex relationships between spectra and target variables \cite{ li_near-infrared_2025, liu_predicting_2026, Pasquini2018NIR,Zhang2022NIRReview}. Among them, regularized linear models such as Ridge regression \cite{hoerl_ridge_1970} provide a simple yet strong baseline by stabilizing estimation in high-dimensional settings, while ensemble methods such as gradient boosting (e.g., Catboost \cite{Prokhorenkova2019CatBoost}) offer flexible nonlinear modeling with strong empirical performance on tabular data. Deep learning models, in particular one-dimensional convolutional neural networks, have also been explored to exploit the ordered structure of spectra, although they typically require careful tuning and sufficient data \cite{nicon_yam, Mishra2022DLNIR}.

~

Despite these advances, a fundamental trade-off remains. Classical chemometric models are attractive because they are simple, well established, and particularly suited to highly collinear, small-sample spectroscopic settings. However, they may remain limited when the relationship between spectra and target variables departs from a predominantly linear structure \cite{limits_linear_nirs}. By contrast, modern machine-learning and deep-learning methods can capture more complex nonlinear patterns, but they often require more extensive hyperparameter tuning and may be more sensitive to dataset size \cite{cat_vs_dl_tabular, dl_nirs_complex_tuning}.

~

A promising direction is offered by Prior-Data Fitted Networks (PFNs), a class of models that have first been introduced by Müller \textit{et al.} in 2021 \cite{muller_transformers_2024}. Their concept relies on pre-training a large transformer, i.e. a neural-network architecture based on self-attention mechanisms for modeling dependencies between input elements \cite{Vaswani2017Attention}, across multiple synthetic datasets to solve classification or regression problems in many contexts. Consequently, unlike classical machine learning approaches, PFNs are calibrated in advance and do not need any parameter optimization when given an unseen dataset. Predictions are performed in a single forward-pass through the pre-trained network. More precisely, both features and targets of the new dataset are taken as input of the pre-trained model. The test targets are masked to avoid data leakage. The model predicts a probability distribution for every test sample (see Supplementary Material for additional details). PFNs quickly showed satisfactory predictive performances for regression and classification tasks, along with major gains in computation time in a wide range of applications.

~ 

PFNs' most notable instance is TabPFN \cite{hollmann_tabpfn_2023, Hollmann2025TabPFN}. TabPFN can be considered as a tabular foundation model, as described by Grinsztajn \textit{et al.} (2026) \cite{grinsztajn_tabpfn-25_2026}. This model is an effective construction of a PFN specialized for real-world tabular data, that is to say for data typical of what one encounters in practice. Indeed, it is specifically designed to handle common problems encountered with real tabular datasets, including uninformative features, categorical features, outliers or missing values. TabPFN can be defined as a pre-trained transformer across around 100 million synthetic datasets, all generated via a distribution of Structural Causal Models. Its most recent published version \cite{grinsztajn_tabpfn-25_2026} achieved state of the art performances, yielding in a few seconds predictions outperforming tree-based models like Catboost \cite{Prokhorenkova2019CatBoost}, and reaching the accuracy of the most advanced ensemble methods \cite{Erickson2020AutoGluon} tuned for 4 hours. Due to its training on small to medium sized synthetic datasets, TabPFN performs especially well on datasets of up to 50,000 samples and 2,000 features.

~

However, it remains unclear whether the statistical priors learned by such tabular foundation models transfer effectively to spectroscopic datasets, which exhibit strong wavelength correlations and domain-specific variability structures. This question is further motivated by the structural differences between NIRS data and typical tabular datasets. In tabular settings, features are usually treated as unordered variables with heterogeneous semantics across columns. In contrast, NIRS spectra exhibit a strong ordering and smoothness along the wavelength axis, with adjacent variables being highly correlated and carrying redundant information. This induces an intricate structure that is not explicitly encoded in standard tabular representations. As a result, it is not obvious whether models designed under generic tabular priors, such as TabPFN, can fully exploit the underlying spectral organization or whether this mismatch may limit their performance.

~

In this study, we investigate this question through a large-scale benchmark study on multiple NIRS datasets covering both regression and classification tasks. TabPFN is compared against representative baselines including classical chemometric models, regularized linear models, tree-based ensemble methods, and neural architectures adapted to spectral data.

~

Beyond this general objective, the study is designed to address the specific challenges of NIRS modeling through a controlled and reproducible experimental framework. First, we construct a large and heterogeneous benchmark gathering a wide range of datasets, covering different sample sizes, dimensional regimes, and application domains. Second, we define a unified evaluation protocol in which preprocessing selection and hyperparameter optimization are performed jointly using cross-validation, while preserving a strict separation with an independent test set. Third, we introduce a structured preprocessing search strategy reflecting the physical and statistical nature of spectral transformations, allowing a fair comparison across models without exhaustive enumeration of all possible pipelines.

~

The contributions of this work are threefold. First, we provide a large-scale evaluation of a tabular foundation model as a calibration engine for NIR chemical sensing data. Second, we quantify how preprocessing affects this model relative to established chemometric and machine-learning baselines. Third, we assess its behavior under practically important deployment conditions, including spectral outliers and extrapolation beyond the calibration domain.

~

These elements collectively aim to bridge the gap between the conceptual promise of tabular foundation models and their effective use in applied spectroscopy.

\section{Materials and Methods} \label{sec:experimental_setup}

\subsection{Datasets} \label{datasets}

\begin{figure}[!htbp]
    \centering
    \includegraphics[width=0.92\textwidth]{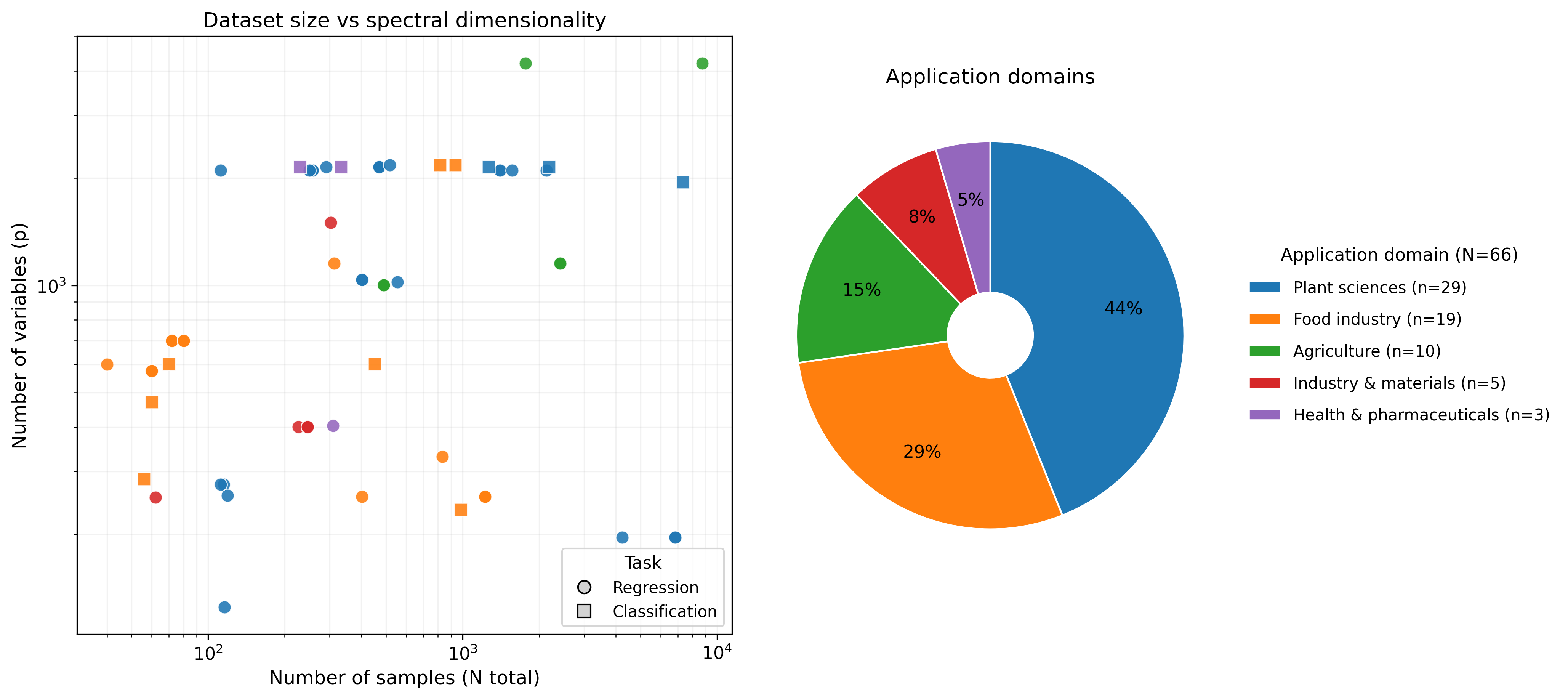}
    \caption{Scatter plot representing the diversity of datasets in sample sizes and number of variables. Pie chart showing the diversity of application domains related to the benchmark.}
    \label{fig:dataset_diversity}
\end{figure}

The benchmark was designed to cover a broad range of NIRS prediction problems representative of current applied chemometric practice. To this end, we assembled a multi-dataset collection spanning heterogeneous analytical contexts, sample types, target variables, and number of features. Fig.~\ref{fig:dataset_diversity} illustrates this diversity in terms of shape and scope. The objective was not to focus on a specific application domain, but rather to evaluate model behaviour across a diversified set of realistic NIRS calibration tasks. The benchmark includes quantitative prediction of chemical or physicochemical properties such as moisture, protein, starch, amylose, oil, carotenoids, dry matter, or related quality traits, as well as classification tasks associated with sample identity, origin, or condition. Several datasets share a common \texttt{database} of spectra, only differing in the associated target variable. 

~

The benchmark is intentionally heterogeneous with respect to sample size and dimensionality, which is particularly important in NIRS. In the present benchmark, the regression collection comprises 54 datasets, with a median size of 402 samples and a median dimensionality of 1003 variables, while the classification collection comprises 12 datasets, with a median size of 876 samples and a median dimensionality of 2151 variables. Across the full benchmark, dataset size ranges from 56 to 8731 samples, and the number of spectral variables ranges from 125 to 4200, reflecting the diversity of NIRS settings encountered in practice. The datasets also cover a wide variety of response scales, levels of between-sample variability, and calibration/test configurations. Such diversity is essential for a fair benchmark, since the relative strengths of chemometric, machine-learning, and foundation-model approaches may vary substantially across dataset regimes.

~

Whenever available, we preserved the train/test split protocol originally associated with each dataset in order to remain consistent with established evaluation practice. Otherwise, deterministic external splits were constructed using a \textit{Sample set Partitioning based on joint X-Y distances} (SPXY) \cite{Galvao2005KS}, or its stratified variant for classification. This design ensures that model comparison is performed under realistic validation settings while keeping the final test set strictly excluded from model and preprocessing selection. 

~

No explicit removal of outliers was performed during dataset construction or model evaluation. All samples were kept in their original form in both calibration and test sets. Spectral outliers are instead identified \emph{a posteriori} on the test sets only, and their impact on model performance was examined through a dedicated robustness analysis (see Section~\ref{sec:robustness}).

~

A complete dataset-by-dataset description of the benchmark is provided in the Supplementary Material (Tables~\ref{tab:dataset_description_long_regression} and~\ref{tab:dataset_description_long_classification}).

\FloatBarrier

\subsection{Compared models}

All experiments involving TabPFN were conducted using TabPFN-2.5 \cite{grinsztajn_tabpfn-25_2026}, the version of the model released in November 2025 by Prior Labs. More specifically, the default checkpoint was used in all runs, which corresponds to the real-data fine-tuned variant of TabPFN-2.5 (internally referred to as \texttt{v2.5-real} in the \texttt{tabpfn} Python package).

~

A meaningful evaluation of TabPFN in NIRS requires comparison against a baseline reflecting the main modeling strategies used in chemometrics and applied machine learning. In particular, the benchmark should cover (i) classical linear approaches, (ii) nonlinear machine learning methods, and (iii) deep learning architectures adapted to spectral data.

~

For regression tasks, we therefore considered four representative predictors: Partial Least Squares (PLS), Ridge regression, CatBoost, and a one-dimensional convolutional neural network (CNN-1D) \cite{nicon_yam}. PLS was included as the main chemometric reference due to its widespread use and robustness in handling collinearity through latent variables. Ridge provides a simple yet strong regularized linear baseline. CatBoost represents nonlinear tree-based ensemble methods, known for their strong performance on tabular data. Finally, CNN-1D captures deep learning approaches specifically adapted to one-dimensional spectral signals by exploiting the ordered structure of wavelengths.

~

For classification tasks, the baseline was adapted to include PLS-DA, the variant of PLS specifically designed for classification, CatBoost, and a classification version of the same CNN-1D architecture. This choice preserves the same logic as in regression while focusing on methods commonly used or readily adaptable to discriminant problems in spectroscopy.

~

In the present benchmark, a CNN-1D architecture was used as the representative of deep learning baselines because it combines two desirable properties for spectroscopy. First, unlike generic multilayer perceptrons, it explicitly exploits the ordered structure of spectral variables. Second, unlike larger deep architectures, it remains relatively lightweight and therefore usable on the small- to medium-sized datasets typically encountered in NIRS. This makes it a meaningful comparison point for TabPFN: both approaches aim to provide strong predictive performance beyond classical linear chemometrics, but they rely on very different learning paradigms. Whereas the CNN-1D must be trained and tuned on each dataset, TabPFN operates through training-free in-context inference once pretrained.

~

All experiments were conducted within the \texttt{nirs4all} framework \cite{beurier2025nirs4all}, an open-source Python package that provides unified tools for NIRS dataset management, preprocessing pipeline construction, model calibration, cross-validation, and evaluation. The CNN-1D architecture used as the deep learning baseline in this benchmark is part of the model collection integrated in this library.

~

This setup allows us to assess whether a foundation model pretrained on large collections of synthetic tabular tasks can compete with, or even surpass the baseline of carefully trained models on the set of NIRS datasets that we gathered.

\subsection{Preprocessing Search Space} \label{preprocs}

Preprocessing is a central component of NIRS modeling, as raw spectra are affected by multiple physical perturbations such as light scattering, baseline drift, and instrumental noise. Rather than detailing each method individually, we consider representative transformations from the main preprocessing families commonly used in chemometrics \cite{Rinnan2009Preprocessing}.

~

Baseline effects are addressed using Asymmetric Least Squares (ASLS) \cite{eilers_baseline_2005}, which corrects slowly varying background distortions.

~

Scatter effects, arising from multiplicative and additive variations in light propagation, are handled using Standard Normal Variate (SNV) \cite{barnes_standard_1989} and Extended Multiplicative Signal Correction (EMSC) \cite{Martens1991EMSC}.

~

Noise reduction and derivative-based feature enhancement are performed using Savitzky--Golay (SG) filters \cite{Savitzky1964SG}, with several configurations combining different window sizes, polynomial orders, and derivative orders. For linear models, an additional Gaussian smoothing option is included.

~

We also consider global transformations aimed at improving signal representation. These include area normalization, Orthogonal Signal Correction (OSC) \cite{wold_orthogonal_1998}, and the Haar transform, which provides a multiresolution decomposition of the signal:
\begin{equation}
x = \sum_{k} \langle x, \psi_k \rangle \psi_k,
\end{equation}
where $(\psi_k)_k$ denotes the Haar wavelet basis.

~

Finally, dimensionality reduction and feature scaling are incorporated depending on the model. For TabPFN and CatBoost, an optional PCA step is applied, retaining $25\%$ of the original features. The $25\%$ threshold was selected as a pragmatic compromise after preliminary experiments showing that variance-based PCA thresholds often retained very few components in highly collinear spectra, occasionally removing low-variance but chemically informative directions. Since the objective was not to optimize PCA itself but to provide a controlled redundancy-reduction option across many datasets, a fixed feature-retention ratio was preferred. This choice is motivated by the strong collinearity of NIRS spectra: variance-based PCA often retains only a few components, potentially discarding predictive but low-variance information. Retaining a fixed proportion of features provides a more balanced trade-off, as supported by preliminary experiments. For linear models, two additional feature scaling methods (StandardScaler and MinMaxScaler) are also considered.

~

Overall, this preprocessing space covers the main families used in NIRS while remaining computationally tractable, enabling a fair comparison across models and datasets.

\subsubsection{Design of preprocessing search spaces}

Rather than exploring all preprocessing combinations exhaustively, we adopt a two-stage hierarchical search strategy that reflects the structure of NIRS signal processing \cite{Rinnan2009Preprocessing}.

~

The first stage focuses on transformations that directly address physical perturbations in the spectra such as baseline drift, scattering and instrumental noise. A Cartesian product of baseline-derivative methods and scatter correction methods is explored. For each model and dataset, configurations are evaluated by cross-validation, and only the top three pipelines are retained.

~

In the second stage, additional transformations are applied to improve predictive performance from a statistical perspective (dimensionality reduction, normalization, feature scaling). This phase operates only on the reduced set of candidates from the first stage, which significantly limits computational cost.

~

This hierarchical design provides three key advantages: (i) it reflects the physical interpretation of preprocessing in spectroscopy, (ii) it avoids exploring redundant or poorly interpretable pipelines, and (iii) it ensures tractability in a large-scale multi-dataset benchmark.

~

In addition, slightly larger search spaces are allocated to linear models (PLS and Ridge). This asymmetric tuning effort reflects practical chemometric usage rather than an attempt to maximize every model equally. Linear models are known to be highly sensitive to preprocessing and latent-variable or regularization choices, whereas TabPFN and CatBoost are commonly used with comparatively robust default settings. Nevertheless, this design implies that the comparison should be interpreted as a realistic workflow-level benchmark rather than as an exhaustive search for the theoretical optimum of each model.

~

The resulting preprocessing search spaces, together with the associated hyperparameter optimization strategies, are summarized in Table~\ref{tab:unified_search_space}.

\subsection{Hyperparameter and Preprocessing Optimization}

For each dataset, the available samples were first divided into a calibration set and an independent test set. The test set was kept strictly untouched throughout model and preprocessing selection.

~

Model selection was performed exclusively on the calibration set using a three-fold cross-validation procedure based on SPXY \cite{Galvao2005KS}. The same validation scheme was used jointly for preprocessing selection and hyperparameter optimization throughout the benchmark. Candidate models were therefore defined as complete pipelines combining preprocessing steps and model hyperparameters.

~

For each complete preprocessing configuration constructed from the predefined search space described above, all candidate hyperparameter configurations were evaluated on the calibration set using the predefined validation splits. The validation scores were averaged across folds, and the hyperparameter configuration yielding the best mean validation performance was retained for this preprocessing configuration.

~

This procedure was repeated for every candidate preprocessing configuration. At the end of the search, each preprocessing configuration was therefore associated with its own best hyperparameter configuration. For each dataset, the final model corresponded to the complete pipeline — including both preprocessing steps and hyperparameters — that achieved the highest mean cross-validation score across folds. Once this optimal pipeline had been identified, the model was retrained on the full calibration set using the selected preprocessing and hyperparameter configuration, and its final predictive performance was evaluated once on the independent test set.

~

This protocol ensures that preprocessing choice and hyperparameter tuning are performed in a fully consistent way within the calibration data only, while preserving a strict separation between model selection and final external evaluation.

~

The hyperparameter optimization strategy was to keep the comparison realistic from an applied NIRS perspective, that is, to tune each method according to procedures consistent with its practical use while keeping the overall benchmark computationally tractable.

~

For PLS, the number of latent components was optimized within the range ${1,\dots,30}$. For each candidate preprocessing pipeline, all candidate values were evaluated under the predefined validation scheme, and the number of latent components minimizing the mean validation RMSE was retained.

~

For Ridge and CNN-1D, hyperparameter optimization was performed with Optuna \cite{akiba_2019_optuna} using a Tree-structured Parzen Estimator (TPE) sampler \cite{bergstra_2011_tpe}, with 30 trials per preprocessing pipeline. TPE is a sequential model-based optimization method that builds probabilistic models of promising and non-promising regions of the search space, and then samples new candidate configurations preferentially in regions expected to yield better objective values. In practice, it provides a more efficient alternative to exhaustive grid search when the search space is continuous or high-dimensional.

~

For TabPFN and CatBoost, no additional hyperparameter optimization was performed beyond the predefined ensemble settings used during model selection and final refitting. This choice was motivated by two considerations. First, the benchmark already involves a large preprocessing search over many datasets, so keeping TabPFN and CatBoost hyperparameters fixed helps maintain a reasonable overall computational cost. Second, these methods are known to achieve strong performance with near-default settings, which is consistent with the results reported later in this study.

~

A practical limitation of the CNN-1D baseline should nevertheless be emphasized. In the present benchmark, this model is tuned within a predefined \texttt{nirs4all} hyperparameter search space, including architectural choices such as kernel sizes, convolutional widths, and related structural parameters (for more details see Supplementary Material). As a consequence, for some of the smallest datasets in the benchmark, these architectural constraints make CNN-1D training unfeasible. In such cases, no exploitable result was obtained for CNN-1D specifically.

~

A compact summary of these model-specific optimization rules, together with the preprocessing search spaces, is provided in Table~\ref{tab:unified_search_space}.

\begin{table}[!htbp]
\centering
\caption{Summary of preprocessing search spaces and hyperparameter optimization strategies for each model. Preprocessing spaces are expressed as Cartesian products of transformation sets, with Phase-2 applied only to the top-3 pipelines retained from Phase-1. \textit{None} means no preprocessing is applied at that step. The total number of configurations corresponds to the number of evaluated preprocessing pipelines, computed from the hierarchical two-phase search (full Phase-1 search followed by Phase-2 expansion of the top-3 retained pipelines), multiplied by the number of hyperparameter configurations and the number of cross-validation folds.}
\label{tab:unified_search_space}
\footnotesize
\renewcommand{\arraystretch}{2.8}
\setlength{\tabcolsep}{4pt}

\begin{tabularx}{\textwidth}{
>{\centering\arraybackslash}p{1.35cm}
>{\raggedright\arraybackslash}X
>{\raggedright\arraybackslash}X
>{\raggedright\arraybackslash}p{2.0cm}
>{\raggedright\arraybackslash}p{1.7cm}
>{\raggedright\arraybackslash}p{2.2cm}
>{\raggedright\arraybackslash}p{1.8cm}}
\toprule
Model & Phase 1: signal correction & Phase 2: representation / scaling & Hyperparameter search & Optimization method & Final training setting & Total configurations per dataset \\
\midrule

\multirow[c]{3}{*}{\centering TabPFN}
& \multirow[c]{3}{=}{\makecell[l]{
$\{\texttt{None},$ \\
$\texttt{ASLSBaseline},$ \\
$\texttt{SavGol(11,2,1)},$ \\
$\texttt{SavGol(15,2,1)},$ \\
$\texttt{SavGol(21,2,1)},$ \\
$\texttt{SavGol(15,3,2)},$ \\
$\texttt{SavGol(21,3,2)}\}$ \\
$\times$ \\
$\{\texttt{None},$ \\
$\texttt{SNV},$ \\
$\texttt{EMSC}\}$
}}
& \multirow[c]{3}{=}{\makecell[l]{
Top-3 Phase-1 \\
$\times$ \\
$\{\texttt{None},$ \\
$\texttt{PCA}_{0.25},$ \\
$\texttt{OSC}\}$
}}
& \rule{0pt}{3.2ex} Fixed hyperparameter \texttt{n\_estimators = 1}.
& Fixed settings
& Refit on full calibration set, \texttt{n\_estimators = 16}.
& $(21 + 3\times3)\times 1 \times 3 = 90$
\\
\cmidrule(lr){1-1}\cmidrule(lr){4-7}

\centering CatBoost
& &
& \rule{0pt}{3.2ex} Fixed hyperparameter \texttt{n\_estimators = 200}.
& Fixed settings
& Refit on full calibration set, \texttt{n\_estimators = 500}.
& $30 \times 1 \times 3 = 90$
\\
\cmidrule(lr){1-1}\cmidrule(lr){4-7}

\centering CNN-1D
& &
& \rule{0pt}{3.2ex} Predefined \texttt{nirs4all} hyperparameter space.
& Optuna-TPE, 30 trials
& Best selected configuration refit on the full calibration set.
& $30 \times 30 \times 3 = 2700$
\\

\midrule

\multirow[c]{2}{*}{\centering PLS}
& \multirow[c]{2}{=}{\makecell[l]{
$\{\texttt{None},$ \\
$\texttt{ASLSBaseline},$ \\
$\texttt{SavGol(11,2,1)},$ \\
$\texttt{SavGol(15,2,1)},$ \\
$\texttt{SavGol(21,2,1)},$ \\
$\texttt{SavGol(15,3,2)},$ \\
$\texttt{SavGol(21,3,2)},$ \\
$\texttt{Gaussian}\}$ \\
$\times$ \\
$\{\texttt{None},$ \\
$\texttt{SNV},$ \\
$\texttt{EMSC}\}$
}}
& \multirow[c]{2}{=}{\makecell[l]{
Top-3 Phase-1 \\
$\times$ \\
$\{\texttt{None},$ \\
$\texttt{Haar},$ \\
$\texttt{AreaNorm},$ \\
$\texttt{OSC}\}$ \\
$\times$ \\
$\{\texttt{None},$ \\
$\texttt{StandardScaler},$ \\
$\texttt{MinMaxScaler}\}$
}}
& \rule{0pt}{3.2ex} \texttt{n\_components} searched exhaustively in $\{1,\dots,30\}$.
& Exhaustive evaluation across folds
& Best \texttt{n\_components} refit on the full calibration set.
& $60 \times 30 \times 3 = 5400$
\\
\cmidrule(lr){1-1}\cmidrule(lr){4-7}

\centering Ridge
& &
& \rule{0pt}{3.2ex} Continuous search of regularization parameter $\alpha$.
& Optuna-TPE, 30 trials
& Best $\alpha$ refit on the full calibration set.
& $60 \times 30 \times 3 = 5400$
\\

\bottomrule
\end{tabularx}
\end{table}

\FloatBarrier
\subsection{Evaluation Metrics}

The predictive performance of the models was assessed at two complementary levels: during cross-validation on the calibration set, where candidate preprocessing pipelines and model configurations were compared, and on the independent test set, where the final selected model was evaluated.

~

For regression tasks, prediction quality was primarily measured using the root mean squared error (RMSE), defined as

\begin{equation}
\operatorname{RMSE}(y,\hat{y})
=
\sqrt{\frac{1}{N}\sum_{k=1}^{N}(y_k-\hat{y}_k)^2},
\end{equation}

where $N$ is the number of predicted samples, $y \in \mathbb{R}^N$ is the vector of observed analytes, and $\hat{y}$ is the corresponding vector of predictions.

~

For classification tasks, predictive performance was assessed using balanced accuracy (ACC), which is particularly appropriate in the presence of class imbalance. In the general multiclass setting, balanced accuracy is defined as the average recall across classes:

\begin{equation}
\mathrm{ACC}
=
\frac{1}{C}
\sum_{c=1}^{C}
\frac{1}{N_c}
\sum_{i \in \mathcal{I}_c}
\mathbf{1}\!\left(\hat{y}_i = y_i\right),
\end{equation}

where $C$ is the number of classes, $\mathcal{I}_c$ denotes the set of samples belonging to class $c$, and $N_c = |\mathcal{I}_c|$.

~

To distinguish the different stages of the whole process, we report three families of metrics. First, during cross-validation, the mean validation performance across folds was used as the model-selection criterion. For regression, this quantity is denoted RMSECV and is defined as

\begin{equation}
\operatorname{RMSECV}
=
\frac{1}{K}\sum_{k=1}^{K}
\operatorname{RMSE}\!\left(
y^{\mathcal{F}_k},
\hat{y}^{\mathcal{F}_k}_{(k)}
\right),
\end{equation}

where $\mathcal{F}_k$ is the $k$-th validation fold and $\hat{y}^{\mathcal{F}_k}_{(k)}$ denotes the predictions obtained from the model trained without fold $\mathcal{F}_k$. The analogous quantity in classification is denoted ACC-CV.

~

Finally, after selecting the optimal preprocessing and hyperparameter configuration, the model was retrained on the full calibration set and evaluated once on the independent test set. The resulting regression metric is denoted RMSEP:

\begin{equation}
\operatorname{RMSEP}
=
\operatorname{RMSE}(y^{\mathrm{test}},\hat{y}^{\mathrm{test}}),
\end{equation}

with ACCP denoting the balanced accuracy evaluated on the test set.

~

In addition to absolute predictive errors, we also considered a relative improvement metric in regression in order to quantify the benefit brought by preprocessing or by a given model with respect to a reference configuration. For this purpose, we used the improvement ratio of RMSEP (iRMSEP), following the general principle recently introduced in the NIRS literature \cite{wang_comprehensive_2025}.

~

For a reference model with prediction error $\mathrm{RMSEP}^{\mathrm{ref}}$ and a compared model with prediction error $\mathrm{RMSEP}^{\mathrm{cmp}}$, iRMSEP is defined as
\begin{equation}
\mathrm{iRMSEP}
=
100 \times
\frac{
\mathrm{RMSEP}^{\mathrm{ref}}-\mathrm{RMSEP}^{\mathrm{cmp}}
}{
\mathrm{RMSEP}^{\mathrm{ref}}
}.
\end{equation}

A positive iRMSEP indicates that the compared model improves upon the reference configuration, whereas a negative value indicates a degradation in predictive performance. In the present benchmark, the reference taken is PLS, since it is the standard chemometric predictor in NIRS. These relative-effect measures are more suitable for model comparisons across datasets with different response scales.

~

Taken together, these metrics allow us to disentangle two complementary aspects of model performance: fold-wise validation behavior during model selection, and final generalization performance after refitting on the full calibration data.

\subsection{Robustness to spectral outliers and extrapolation} \label{sec:robustness}

\subsubsection{Spectral-outlier evaluation}
In addition to global predictive performance, we assessed model robustness to spectrally atypical samples. For each dataset, spectral outliers were identified only in the test set, without removing them from the calibration or evaluation procedure. More precisely, spectra were first transformed by SNV, then a PCA model was fitted on the calibration set only. The number of retained components was chosen as the smallest number explaining at least $95\%$ of the calibration variance. Test spectra were projected into this PCA space and their Hotelling's $T^2$ statistic \cite{hotelling_t2} was computed. Samples whose $T^2$ exceeded the $95\%$ threshold were considered spectral outliers. Let $\mathcal{I}_{\mathrm{out}}$ denote the corresponding set of test indices. Model performance on this subset was quantified by
\[
\mathrm{RMSEP}_{\mathrm{out}}
=
\operatorname{RMSE}\!\left(
y^{\mathrm{test}}_{\mathcal{I}_{\mathrm{out}}},
\hat{y}^{\mathrm{test}}_{\mathcal{I}_{\mathrm{out}}}
\right)
\]
Relative performance was then expressed using iRMSEP with PLS as reference.

~

\subsubsection{Extrapolation evaluation}
To assess extrapolation ability, we identified test samples whose target values fall outside the range observed in the calibration set. Formally, letting $[y_{\min}^{\mathrm{train}}, y_{\max}^{\mathrm{train}}]$ denote the range of $Y^{\mathrm{train}}$, we define
\[
\mathcal{I}_{\mathrm{extra}} =
\left\{
i : y_i^{\mathrm{test}} < y_{\min}^{\mathrm{train}}
\ \text{or}\
y_i^{\mathrm{test}} > y_{\max}^{\mathrm{train}}
\right\}
\]
When $\mathcal{I}_{\mathrm{extra}} \neq \emptyset$, RMSE and iRMSEP were computed on this subset.

\subsection{Statistical analysis of model performance}

In order to assess whether the observed differences in predictive performance between models are statistically significant across the benchmark, we adopted a statistical analysis framework commonly used in multi-dataset machine learning studies. This framework combines rank-based global tests, and post-hoc pairwise comparisons.

~

For the statistical comparison based on ranks, aggregation was performed at the \emph{database level}. Let $\mathcal{B}=\{1,\dots,B\}$ denote the set of databases, and let $\mathcal{D}_b$ denote the set of datasets belonging to database $b$. For a given model $j$, let $s_{i,j}$ denote its predictive score on dataset $i$, where lower values indicate better performance in regression (RMSEP) and higher values indicate better performance in classification (ACCP). For each database $b$ and model $j$, we first compute an aggregated performance score
\[
\bar{s}_{b,j} =
\frac{1}{|\mathcal{D}_b|}
\sum_{i \in \mathcal{D}_b} s_{i,j},
\]
that is, the mean predictive performance of model $j$ across all datasets belonging to database $b$.

~

Models are then ranked according to these aggregated scores, with rank~1 assigned to the best-performing model. Let $r_{b,j}$ denote the rank obtained by model $j$ on database $b$. The average rank of model $j$ across the $B$ databases is defined as
\[
R_j = \frac{1}{B}\sum_{b=1}^{B} r_{b,j}.
\]

~

This aggregation strategy ensures that databases contributing many datasets do not dominate the rank-based comparison solely because of their larger internal multiplicity. Instead, each database contributes one rank vector to the global statistical analysis, regardless of how many datasets it contains.

~

To determine whether the observed rank differences are statistically significant, we applied the Friedman test \cite{Friedman_test}, a non-parametric alternative to repeated-measures ANOVA that does not assume normality of the performance distributions. The null hypothesis of the Friedman test states that all models perform equivalently across databases, that is,
\[
H_0 : R_1 = R_2 = \dots = R_k,
\]
where $k$ is the number of compared models. When the null hypothesis is rejected, it indicates that at least one model performs significantly differently from the others at the database level.

~

When the Friedman test indicated a significant global difference, we performed a Nemenyi post-hoc test \cite{nemenyi_posthoc_test} in order to identify which pairs of models differed significantly. The Nemenyi procedure compares the difference between the average ranks of two models $j_1$ and $j_2$ to a critical distance
\[
CD = q_{\alpha} \sqrt{\frac{k(k+1)}{6B}},
\]
where $q_{\alpha}$ is a critical value derived from the Studentized range distribution, $k$ is the number of models, and $B$ is the number of databases. Two models are considered significantly different if the absolute difference between their average ranks exceeds this critical distance. The results are summarized using critical difference diagrams.

\subsection{Computing hardware and environment}

All experiments were conducted on a workstation running Ubuntu Linux. The system was equipped with an Intel Core i9-14900KF CPU (24 cores, 32 threads) and 192\,GB of RAM. Computations involving CatBoost and the CNN-1D model were accelerated using an NVIDIA GeForce RTX 5080 GPU with 16\,GB of VRAM and CUDA 12.9 support. For technical reasons, the GPU-enabled version of TabPFN could not be used in this study, and all TabPFN experiments were therefore performed on CPU. CPU-based computations were parallelized during cross-validation using up to 18 concurrent threads.

~

The software environment was based on Python 3.12.3, with PyTorch 2.11.0 (CUDA-enabled), scikit-learn 1.6.1, and Optuna 4.4.0 as the main libraries used for model development and optimization. All computations were performed in a controlled environment with fixed random seeds and deterministic settings when supported by the underlying libraries.

\section{Results} \label{sec:results}

\subsection{Regression Results}

We first focus on the performances for the regression task, across 54 different datasets. The predictive performances will be examined, along with the influence of preprocessing on the results and the robustness of each model to outliers or extrapolated values. In the following, TabPFN-Raw denotes the direct application of TabPFN to raw spectra, without any preprocessing step, whereas TabPFN-opt denotes the same pretrained TabPFN model embedded in the proposed preprocessing-selection framework, in which the preprocessing pipeline is optimized on the calibration set. A complete table of datasetwise performances is provided in the Supplementary Material.

\subsubsection{Predictive performance across datasets}

\begin{figure}[!htbp]
    \centering
    \includegraphics[width=0.92\textwidth]{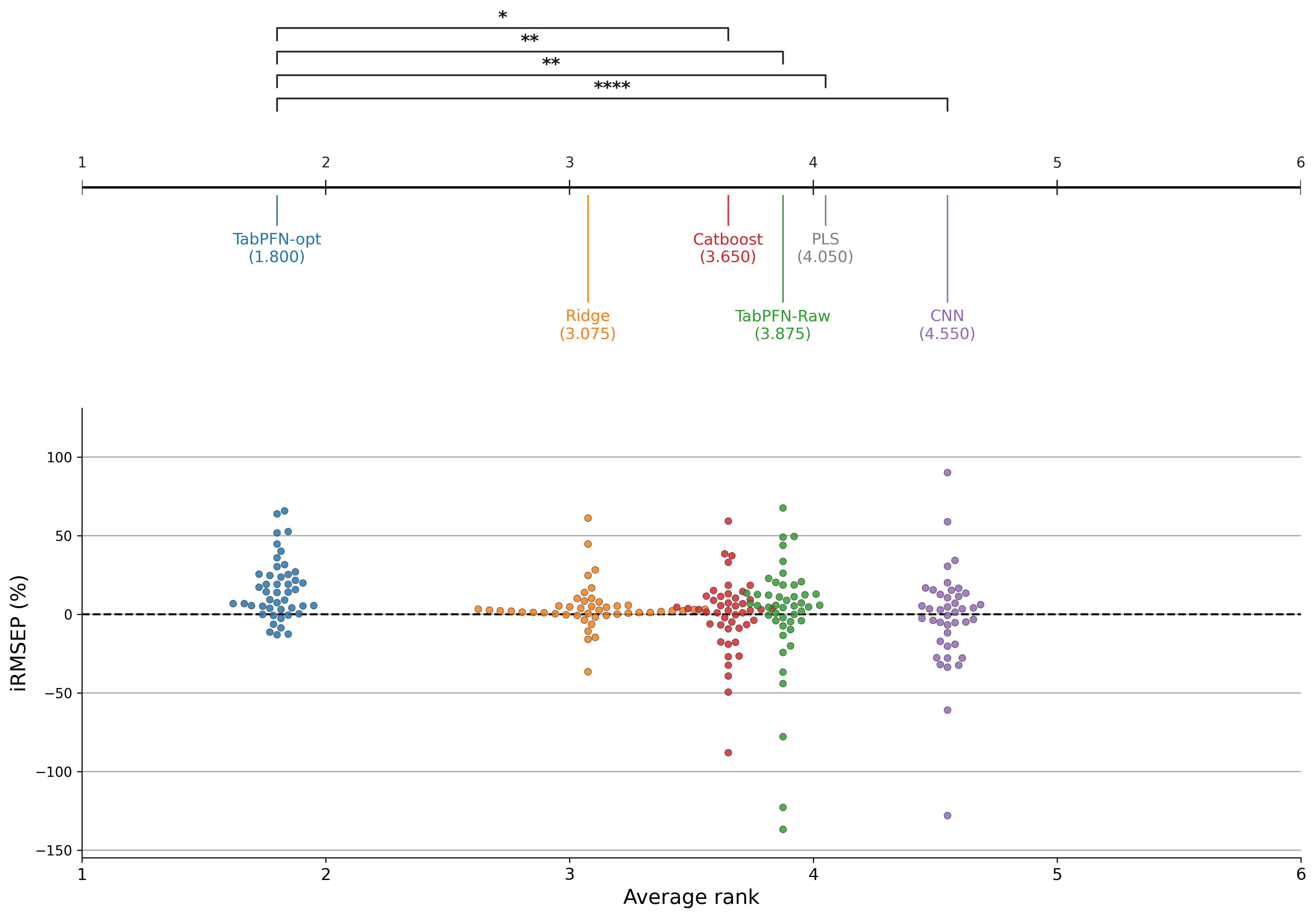}
    \caption{Critical difference diagram and beeswarm plot of iRMSEP values relative to PLS across regression datasets. Positive values indicate improved performance over PLS. For visual clarity, the beeswarm excludes a small number of extreme dataset-level iRMSEP values identified by a conservative 10$\times$IQR (interquartile range) display filter (excluding at the same time all iRMSEP values from the same dataset). The critical difference diagram is computed from the common subset of databases for which all compared models provide valid results. Full dataset-level values are given and displayed in the Supplementary Material, Table~S\ref{tab:complete_results_reg} and  Fig.~S\ref{fig:heatmap_regression_irmsep_vs_pls}.}
    \label{fig:global_irmsep_beeswarm}
\end{figure}

Fig.~\ref{fig:global_irmsep_beeswarm} summarizes regression performance through two complementary views: dataset-level iRMSEP distributions relative to PLS, and database-level average ranks derived from the critical-difference analysis. The Friedman test indicates a significant difference between models ($p = 5.9 \times 10^{-5}$, $CD = 1.520$). TabPFN-opt achieves the best average rank (1.800), followed by Ridge (3.075), Catboost (3.650), TabPFN-Raw (3.875), PLS (4.050), and CNN-1D (4.550).

~

At the dataset level, TabPFN-opt shows a predominance of positive iRMSEP values, indicating improvements over PLS across a majority of datasets. Ridge also exhibits a stable distribution centered near zero, reflecting performance close to PLS. In contrast, TabPFN-Raw, Catboost, and CNN-1D display more dispersed distributions. In particular, CNN-1D shows noticeable improvements as well as major degradations compared to PLS. These trends are consistent with the pairwise statistical analysis (Supplementary Table~S\ref{tab:cd_stats_full}), where TabPFN-opt differs significantly from Catboost, TabPFN-Raw, PLS, and CNN-1D, but not from Ridge. The beeswarm plots must be interpreted carefully since some points have been removed to improve the clarity of the figure. For complete results, see the tables in the Supplementary Material.

~

Fig.~\ref{fig:cumulative_irmsep_dataset_size} provides a complementary view by reporting cumulative iRMSEP values across datasets ordered by increasing sample size. TabPFN-opt remains above PLS across the full range of dataset sizes, while Ridge follows a flatter trajectory with occasional improvements. TabPFN-Raw exhibits cumulative improvements that are similar to its preprocessed version, with several negative iRMSEP in the medium-sized datasets regime. CNN-1D exhibits a more favorable trend on larger datasets, although it remains globally below PLS for the rest of the datasets in this benchmark. A more exhaustive figure is provided in the Supplementary Material, showing cumulative gains depending on the number of variables, and on the number of cells in the dataset (i.e. the number of samples multiplied by the number of variables).

~

\begin{figure}[!htbp]
    \centering
    \includegraphics[width=0.92\textwidth]{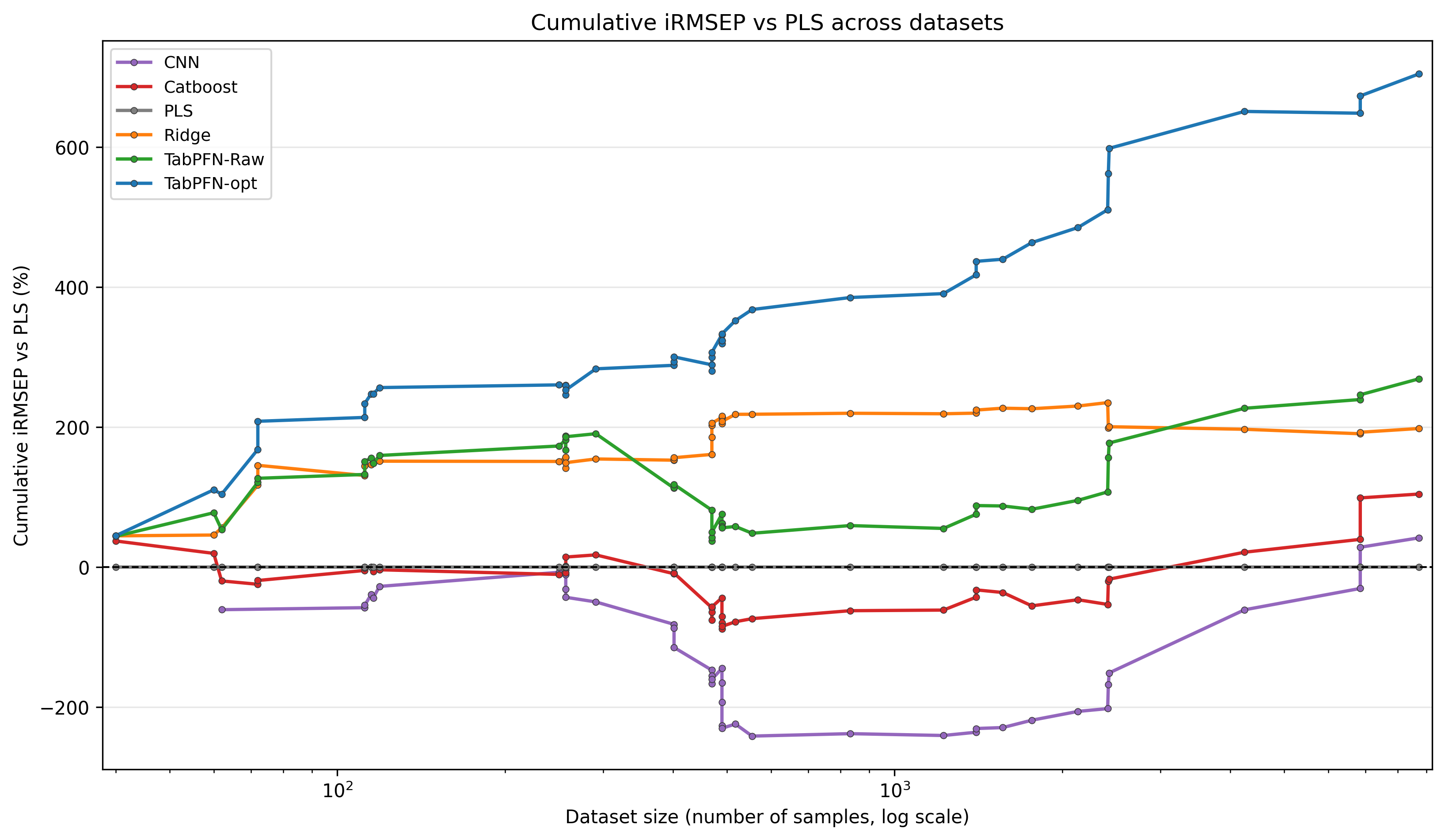}
    \caption{Cumulative iRMSEP versus PLS across regression datasets ordered by dataset sample size. For visual clarity, the visualization excludes a small number of extreme dataset-level iRMSEP values identified by a 10$\times$IQR display filter.}
    \label{fig:cumulative_irmsep_dataset_size}
\end{figure}

Finally, Table~\ref{tab:wtl_regression} reports database-level win/loss counts. TabPFN-opt dominates all competing methods, with clear margins against Catboost (21 wins vs. 3 losses), PLS (18 vs. 4), and Ridge (18 vs. 4), and strongly outperforms TabPFN-Raw (22 vs. 2).

\begin{table}[!htbp]
\centering
\caption{Pairwise win/loss summary on regression databases.}
\label{tab:wtl_regression}
\small
\begin{tabular}{lllrrrrr}
\toprule
Model A & Model B & Wins & Ties & Losses & Win rate & Non-loss rate \\
\midrule
TabPFN-Raw & CNN-1D & 13 & 1 & 7 & 0.650 & 0.667 \\
TabPFN-Raw & Catboost & 16 & 0 & 8 & 0.667 & 0.667 \\
TabPFN-Raw & PLS & 13 & 1 & 9 & 0.591 & 0.609 \\
TabPFN-Raw & Ridge & 8 & 1 & 14 & 0.363 & 0.391 \\

TabPFN-opt & CNN-1D & 17 & 1 & 3 & 0.850 & 0.857 \\
TabPFN-opt & Catboost & 21 & 0 & 3 & 0.875 & 0.875 \\
TabPFN-opt & PLS & 18 & 1 & 4 & 0.818 & 0.826 \\
TabPFN-opt & Ridge & 18 & 1 & 4 & 0.818 & 0.826 \\
TabPFN-opt & TabPFN-Raw & 22 & 1 & 2 & 0.917 & 0.920 \\
\bottomrule
\end{tabular}
\end{table}

\FloatBarrier

\subsubsection{Influence of preprocessing}

\begin{figure}[!htbp]
    \centering
    \includegraphics[width=0.92\textwidth]{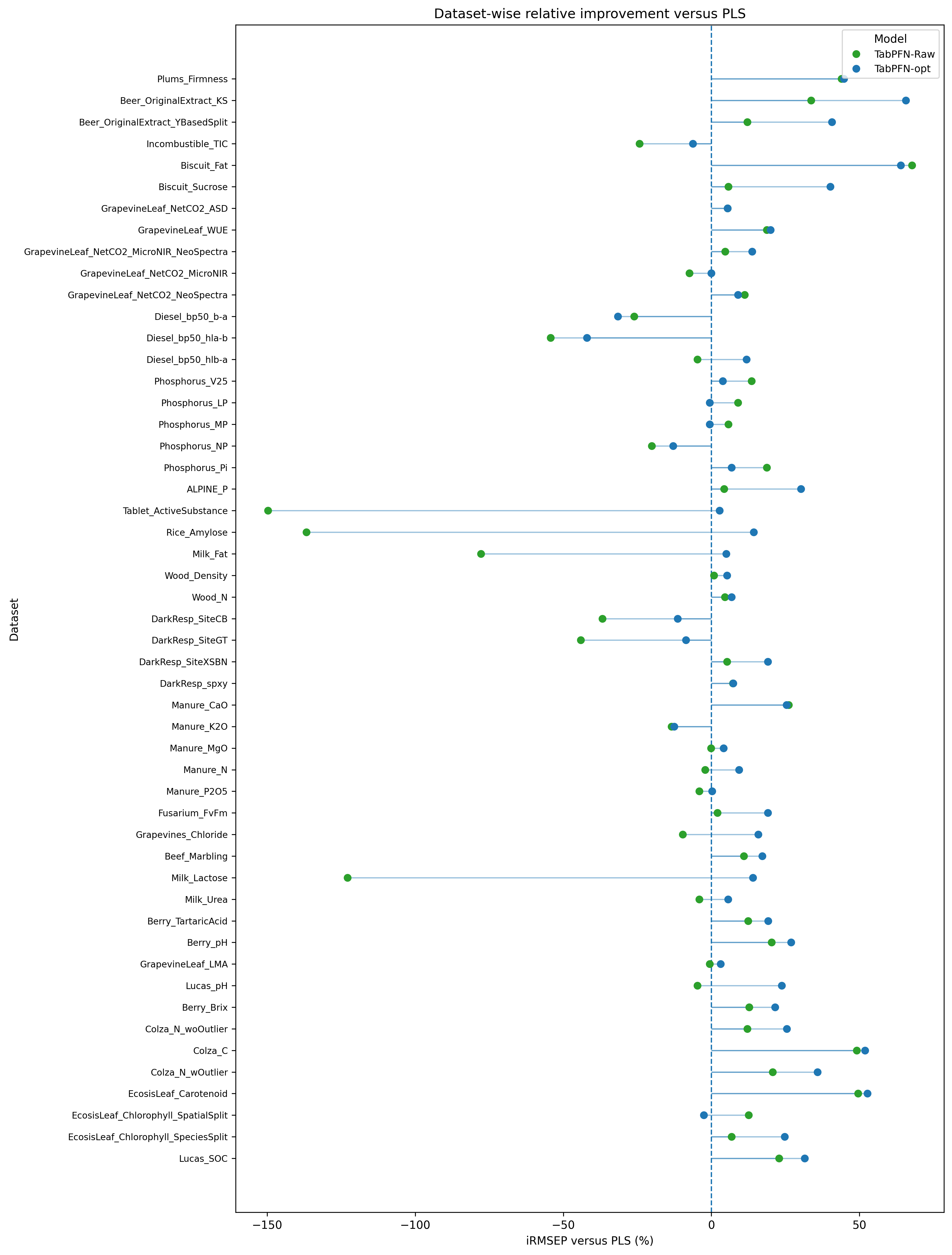}
    \caption{Dataset-wise comparison between TabPFN-Raw and TabPFN-opt in terms of iRMSEP relative to PLS.}
    \label{fig:dumbbell_tabpfn_preprocessing}
\end{figure}

Preprocessing has a variable impact across models and datasets. Fig.~\ref{fig:dumbbell_tabpfn_preprocessing} shows that TabPFN-opt generally improves over TabPFN-Raw, although the magnitude of improvement depends on the dataset. TabPFN-opt being outperformed by TabPFN-Raw means that, on this specific dataset, the best preprocessing configuration is overestimated during the cross-validation procedure and leads to suboptimal predictions on the test set. In such cases, the performances of TabPFN-opt are close to TabPFN-Raw. On the contrary, the optimized version significantly outperforms the raw version on several datasets.

~

The distribution of selected preprocessing components (Figs.~S\ref{fig:preprocessing_linear_models} and S\ref{fig:preprocessing_tabular_models} in the Supplementary Material) highlights different behaviors across model families. Nonlinear models (TabPFN, Catboost, CNN-1D) more often operate with Savitzky-Golay transformations. In particular, TabPFN seems to benefit from smoothing and derivatives with no additional transformation. Linear models (PLS and Ridge) along with CNN-1D frequently select scatter corrections such as SNV. Ridge and Catboost regularly apply signal transformations such as OSC.

~

No single preprocessing pipeline dominates across datasets. While SNV and Savitzky--Golay filtering are frequently selected, their use remains dataset-dependent.

\FloatBarrier

\subsubsection{Robustness to spectral outliers and extrapolation}

\begin{figure}[!htbp]
    \centering
    \includegraphics[width=0.92\textwidth]{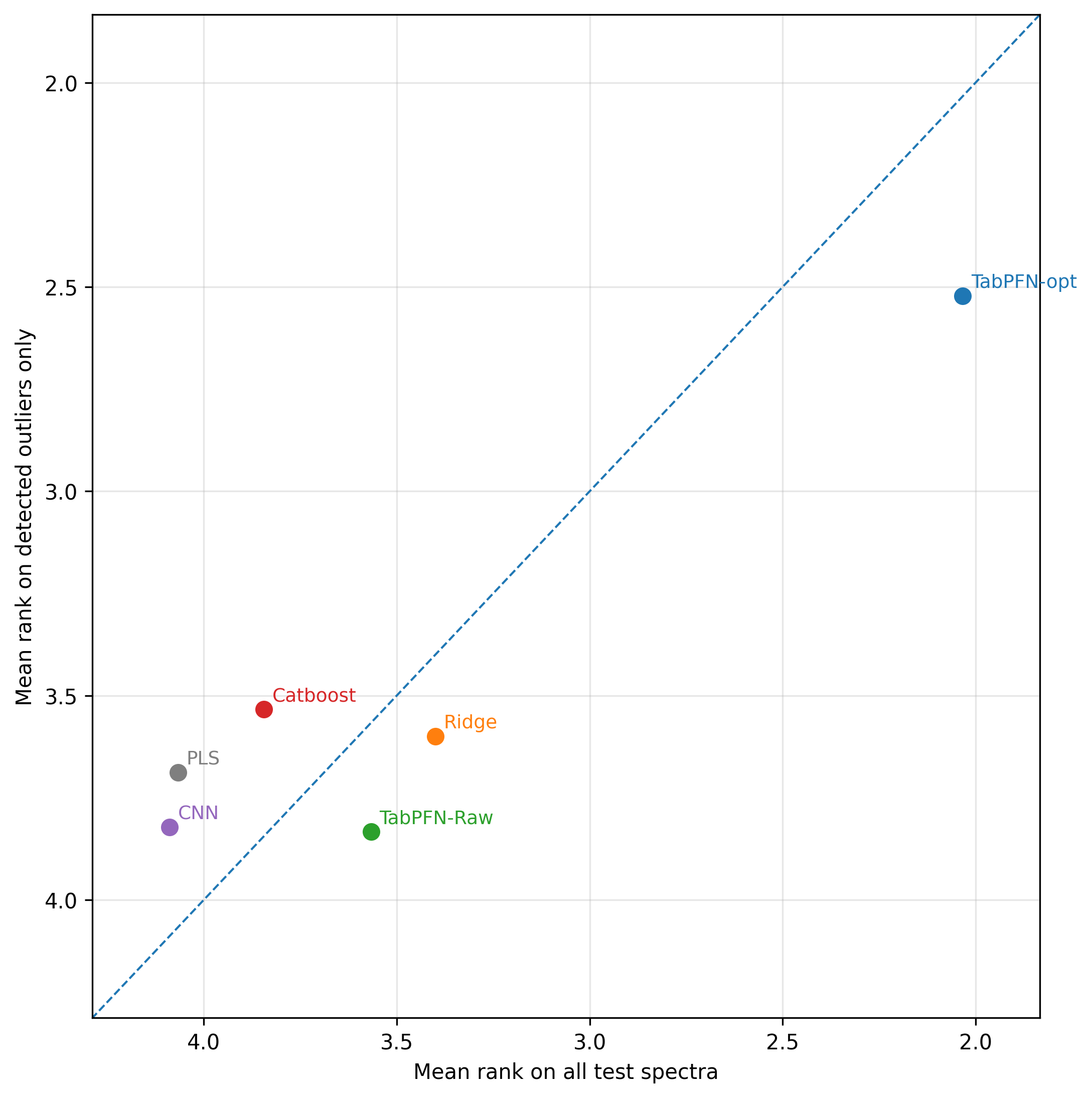}
    \caption{Scatter plot representing the database-wise average rank of each model on all test samples (x axis), versus on test samples that are detected as outliers.}
    \label{fig:robustness_scatter_outlier_ranks}
\end{figure}

Spectral outliers were identified in 48 out of 54 regression datasets. Fig.~\ref{fig:robustness_scatter_outlier_ranks} compares model ranks computed on all test samples versus on outlier subsets. Models above the diagonal improve their relative ranking on outliers, while those below the diagonal deteriorate.

~

PLS, Catboost, and CNN-1D exhibit slight improvements in rank on outliers, whereas TabPFN-opt, Ridge, and TabPFN-Raw show moderate decreases. TabPFN-opt remains the best-performing model overall, but its relative advantage is reduced on these subsets.

~

\begin{figure}[!htbp]
    \centering
    \includegraphics[width=0.92\textwidth]{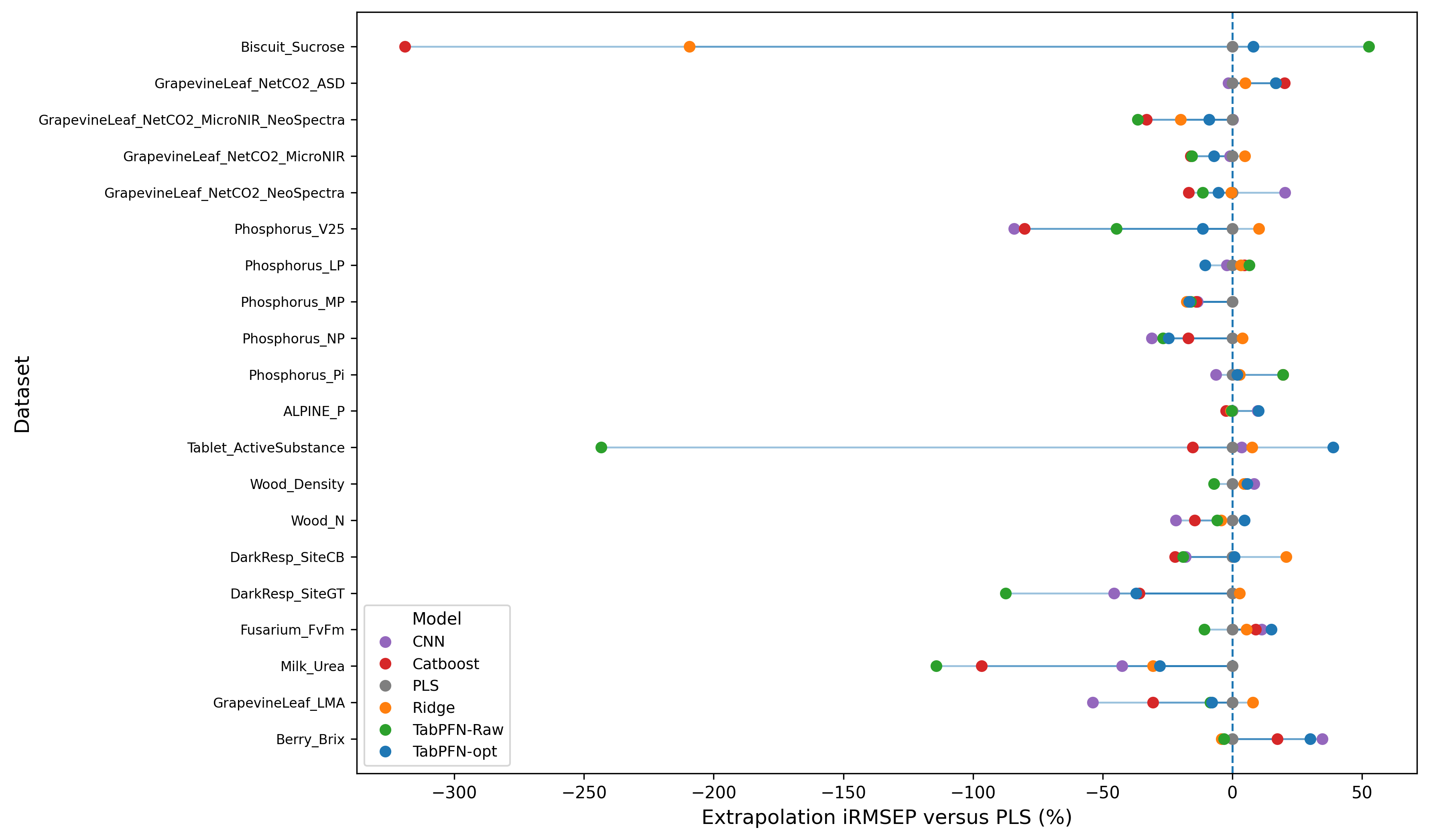}
    \caption{Dataset-wise iRMSEP versus PLS on extrapolated target values (outside the training range). For visual clarity, the visualization excludes extreme dataset-level iRMSEP values identified by a 10$\times$IQR display filter.}
    \label{fig:dumbbell_extrapolation_irmsep}
\end{figure}

Extrapolation regimes were observed in 23 datasets. As shown in Fig.~\ref{fig:dumbbell_extrapolation_irmsep}, all models experience a degradation in performance relative to PLS. TabPFN-opt and Ridge remain closest to PLS, while TabPFN-Raw, Catboost, and CNN-1D show larger deviations.

\FloatBarrier

\subsection{Classification results}

Classification performance was evaluated on 12 datasets using balanced accuracy. As in regression, TabPFN-Raw corresponds to direct inference on raw spectra, while TabPFN-opt includes preprocessing optimization. A complete table of datasetwise performances is provided in the Supplementary Material.

~

\begin{figure}[!htbp]
    \centering
    \includegraphics[width=0.92\textwidth]{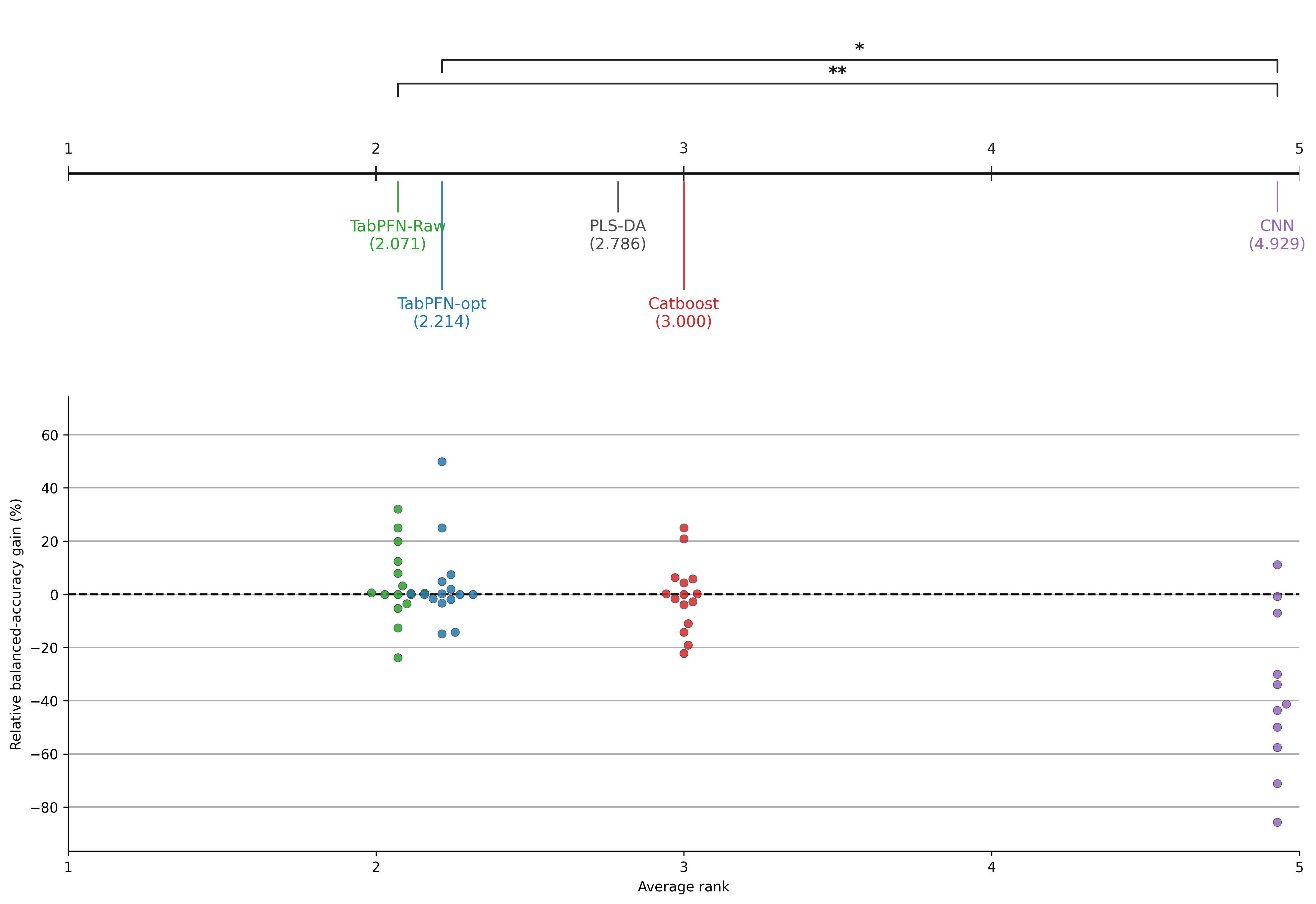}
    \caption{Critical difference diagram and beeswarm plot of relative balanced-accuracy gains versus PLS-DA across classification datasets. Positive values indicate improved performance over PLS-DA. The critical-difference analysis is computed on the common subset of databases for which all compared models provide valid results, including CNN-1D.}
    \label{fig:classif_beeswarm_main}
\end{figure}

Fig.~\ref{fig:classif_beeswarm_main} combines dataset-level accuracy gains relative to PLS-DA with a database-level critical-difference diagram. TabPFN-Raw achieves the best average rank (2.000), followed by TabPFN-opt (2.417), PLS-DA (2.583), Catboost (3.083), and CNN-1D (4.917). The Friedman test indicates a significant global difference ($p = 0.011$, $CD = 2.345$), with TabPFN variants significantly outperforming CNN-1D. At the dataset level, both TabPFN variants show frequent positive gains relative to PLS-DA. Their performance distributions largely overlap, with only a slight advantage for TabPFN-Raw. Catboost shows intermediate behavior, while CNN-1D exhibits more frequent degradations relative to PLS-DA.

~

\begin{figure}[!htbp]
    \centering
    \includegraphics[width=0.92\textwidth]{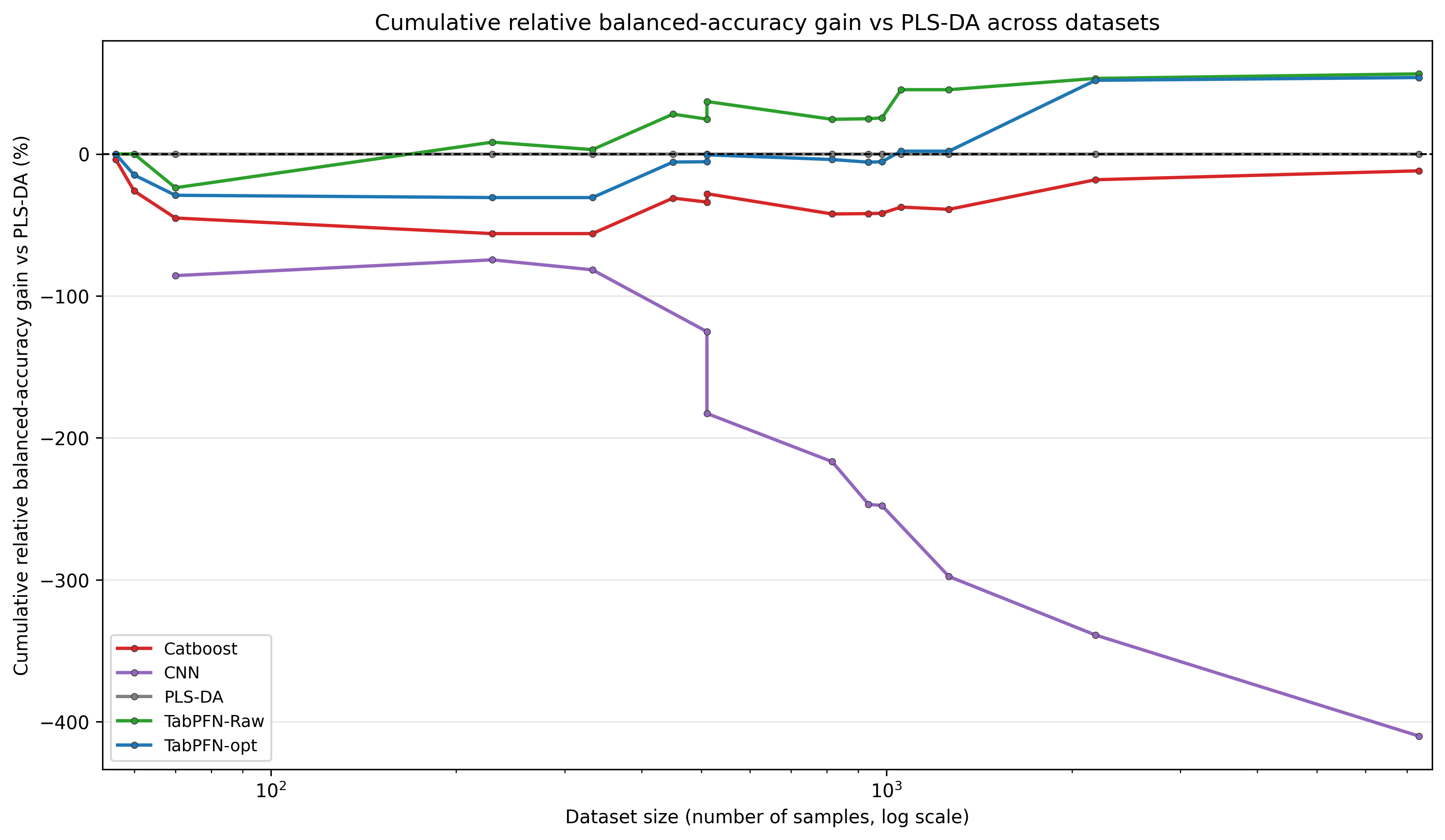}
    \caption{Cumulative relative balanced-accuracy gain versus PLS-DA across classification datasets ordered by dataset size.}
    \label{fig:cumulative_acc_gain_classification}
\end{figure}

\begin{figure}[p]
    \centering
    \includegraphics[width=0.92\textwidth]{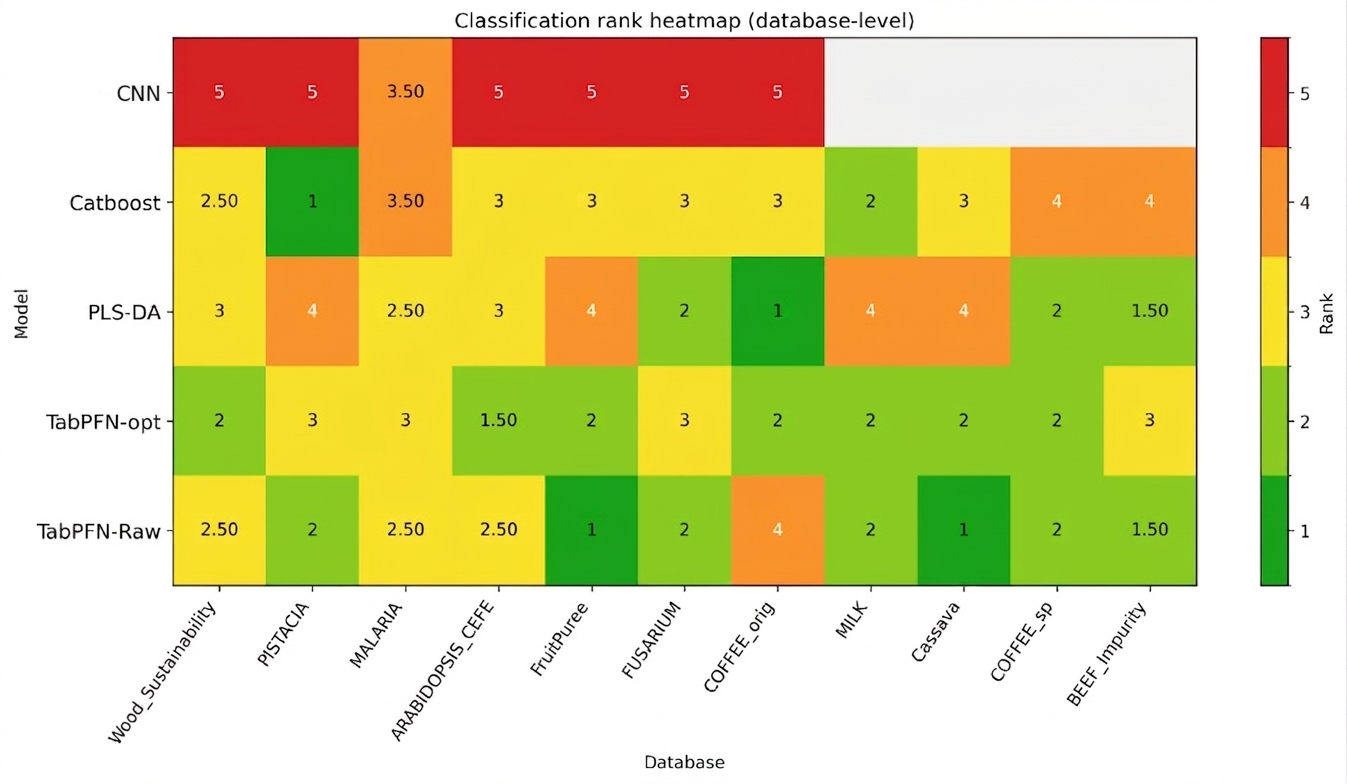}
    \caption{Classification rank heatmap after aggregation at the database level. Lower ranks indicate better predictive performance within each database. Grey cells correspond to unavailable model--database combinations.}
    \label{fig:classif_rank_heatmap}
\end{figure}

Fig.~\ref{fig:cumulative_acc_gain_classification} confirms these trends, with both TabPFN variants accumulating the largest gains across datasets. Dataset-level analyses in the Supplementary Material (Figs.~S\ref{fig:classif_dumbbell}, S\ref{fig:classif_rank_heatmap}) are consistent with these observations.

~

Due to the limited number of datasets, these results should be interpreted cautiously.

\section{Discussion} \label{sec:discussion}

\subsection{Overall behavior across models}

The benchmark highlights that no single model dominates uniformly across all NIRS datasets. Instead, different modeling strategies exhibit distinct performance profiles depending on dataset characteristics.

~

TabPFN-opt emerges as the strongest overall performer in regression, combining favorable average ranks with consistent improvements over the reference PLS model. However, the results also show that simpler approaches remain highly competitive. In particular, Ridge often achieves performance close to PLS and, in some cases, approaches that of TabPFN-opt, indicating that regularized linear models remain strong baselines in high-dimensional, collinear settings.

\subsection{Role of preprocessing for TabPFN}

A key result of this study lies in the comparison between TabPFN-Raw and TabPFN-opt. This comparison isolates the contribution of preprocessing around the same pretrained model.

~

In regression, preprocessing optimization leads to systematic improvements, as reflected by the consistent shift from TabPFN-Raw to TabPFN-opt. This indicates that, although TabPFN can operate directly on raw spectra, part of the spectroscopic variability still benefits from explicit correction steps. The optimal preprocessing configurations obtained across datasets suggest that a major part of this improvement comes from applying a smoothing-derivative operator such as Savitzky-Golay. A careful tuning of this method's parameters, combined with TabPFN, is likely to yield competitive predictions on any NIRS dataset.

~

In contrast, in classification, the two variants perform similarly, suggesting that TabPFN is already able to extract relevant discriminant information from raw spectra. This difference between regression and classification highlights that the role of preprocessing is task-dependent.

~

More generally, these results show that TabPFN does not replace preprocessing in NIRS workflows, but can instead be effectively integrated within them.

\subsection{Dataset regimes and robustness}

The cumulative analyses indicate that the relative performance of TabPFN-opt is not restricted to a specific range of dataset sizes. Its gains over PLS are observed across small, medium, and larger datasets, suggesting that the learned prior transfers across different sample-size regimes.

~

However, the robustness analysis reveals a more nuanced picture. On spectral outliers and extrapolation subsets, the relative advantage of TabPFN-opt decreases, while PLS and, to a lesser extent, Catboost and CNN-1D show improved relative rankings. This suggests that models achieving the best average predictive performance are not necessarily those that are most robust in challenging regions of the data space.

~

These observations reinforce the idea that robustness and peak performance represent complementary properties, both of which are important in applied spectroscopy.

\subsection{Comparison with CNN-1D}

The comparison with CNN-1D illustrates a fundamental difference in learning paradigms rather than an inherent limitation of deep learning for spectral data. CNN-1D must be trained and tuned from scratch on each dataset, which makes its performance more sensitive to the available sample size and the allocated optimization budget. In the present benchmark, all models were compared under a shared and intentionally constrained optimization budget, so that no method was pushed to its theoretical optimum. Under these conditions, CNN-1D shows more variable results across datasets, with a tendency to underperform on smaller calibration sets. However, this should not be interpreted as a general weakness of convolutional architectures for NIRS: with a larger tuning budget, extended data augmentation strategies, or architecture search specifically tailored to spectral data, CNN-1D performance could likely improve substantially. The results reported here reflect a controlled workflow-level comparison, not a definitive ranking of the methods at their full potential.

~

In contrast, TabPFN provides competitive performance without dataset-specific training, which gives it a structural advantage under tight computational budgets. This asymmetry in optimization cost is itself a practically relevant dimension of the comparison.

\subsection{Limitations of the benchmark}

Several limitations should be kept in mind when interpreting these results. The benchmark was designed to remain computationally feasible across a large number of datasets, which necessarily constrained the extent of hyperparameter optimization for all methods. In particular, TabPFN and CatBoost were run with fixed hyperparameters, while CNN-1D was restricted to 30 architectural configurations per preprocessing pipeline and evaluated without any feature augmentation strategy, even though such techniques may constitute an important source of improvement for deep learning approaches on spectral data. Linear models such as PLS and Ridge were, by contrast, evaluated over substantially larger preprocessing and hyperparameter search spaces, reflecting both common chemometric practice and their stronger sensitivity to these choices. This asymmetry implies that the reported performances should not be interpreted as upper bounds: a benchmark allocating larger and more balanced optimization budgets could yield different absolute levels and potentially different relative rankings. In particular, the present study likely provides a conservative estimate of the achievable performance of TabPFN in NIRS, since it was not explored under the full range of configurations that could further improve its predictions.

~

A second limitation concerns the classification benchmark. Only 12 datasets were used, which remains too limited to claim strong generality with respect to the full diversity of NIRS classification problems. The classification results are therefore informative and suggestive, but they should be interpreted with more caution than the regression results. Expanding the number and diversity of classification datasets would be an important next step for consolidating the conclusions drawn here.

\section{Conclusions}

This study evaluated the performance of TabPFN on a large and heterogeneous NIRS benchmark, and compared it with classical chemometric, machine learning, and deep learning approaches.

~

In regression, TabPFN combined with preprocessing optimization (TabPFN-opt) achieved the strongest overall performance, with consistent improvements over the reference PLS model across datasets. In classification, both TabPFN variants performed similarly, indicating that competitive results can already be obtained without preprocessing optimization.

~

At the same time, the results confirm that classical methods remain highly relevant. PLS continues to provide a robust reference across datasets and challenging regimes, while Ridge emerges as a particularly strong and stable baseline.

~

These conclusions should nevertheless be interpreted in light of the benchmark design. The comparison was conducted under a shared and intentionally limited optimization budget, ensuring a fair and reproducible evaluation across a large number of datasets, but at the cost of leaving each method below its full potential. This applies equally to TabPFN, CatBoost, and CNN-1D: none of these models was exhaustively tuned. The present results should therefore be viewed as a robust large-scale comparison under controlled conditions, not as a definitive estimate of what each method can achieve given unlimited computational resources.

~

Overall, TabPFN appears as a powerful complementary tool for NIRS modeling, particularly in small- to medium-sized datasets. Several perspectives emerge from this work. First, TabPFN could likely benefit from a more extensive hyperparameter optimization strategy than the one considered here. Second, the preprocessing framework around TabPFN, especially for regression, could be further improved, in particular through more tailored dimensionality-reduction approaches, which appear especially relevant for highly collinear NIRS data. More broadly, the potential of PFN-based models has only been introduced and partially explored in this study. An important next step would be to investigate priors or pretraining strategies specifically constrained toward NIRS-type data, in order to better capture the structural properties of spectra and further improve domain-specific performance. Rather than replacing existing chemometric approaches, this study extends the range of effective modeling strategies available in practical spectroscopic workflows.

\bibliographystyle{abbrv}
\bibliography{NIRS,ML_Models,TabPFN,Statistical_tests}

\clearpage
\section{Supplementary Material}

\subsection{Theoretical foundations of PFNs} \label{theory_pfn}

\subsubsection{Statistical context}

Let $X \in \mathcal{X} \subset \mathbb{R}^d, \, Y \in \mathcal{Y}$ be a random vector of features and a random vector of targets, respectively.

~ 

Let us consider a classification task: the objective is to find, for any realization $(x,y)$ of $(X,Y)$, the probability $p_0(y|\, x) = \mathbb{P}(Y=y | \, X=x)$.

~ 

In the case of regression, it is in practice transposed to a classification problem by setting bins. Given $\{I_1, I_2, \dots, I_K \}$ a partition of $\mathcal{Y} \subset \mathbb{R}$, then we aim to find the probabilities $\mathbb{P}(Y \in I_k | \, X=x)$ for every $k \in \{1,2,\dots,K\}$.

\subsubsection{Posterior Predictive Distribution}

From a Bayesian nonparametric perspective, $p_0$ is the realization of a random infinite-dimensional model $p \in \mathcal{P}$ following a distribution $\Pi$. $\Pi$ is called the prior, which is a probability distribution induced by all the beliefs we have a priori, that is, before seeing the data.

~

To estimate $p_0$, the following process has been set up:

\begin{enumerate}[label=\roman*]
    \item Draw a model $p$ from the prior $\Pi$;
    \item Draw iid samples $\mathcal{D}_n = (X_i,\, Y_i)_{i=1}^n$ from model $p$;
    \item Draw iid test samples $(X^{\text{test}}, \,  Y^{\text{test}}) = (X_t^{\text{test}}, \,  Y_t^{\text{test}})_{t=1}^T$ from model $p$.
\end{enumerate}

The objective is then to estimate the probabilities: $p_0(y^{\text{test}}|\, x^{\text{test}}) = \mathbb{P}(Y^{\text{test}} = y^{\text{test}}|\, X^{\text{test}} = x^{\text{test}})$.

~

For every $n$, the tuple $(\mathcal{D}_n \cup (X^{\text{test}}, \,  Y^{\text{test}}),\, p)$ defines a joint distribution. An interesting estimation of $p_0(y^{\text{test}}|\, x^{\text{test}})$ is given by Posterior Predictive Distributions (PPD), defined as follows for every $n$:

$$\pi(y^{\text{test}}|\, x^{\text{test}}, \, \mathcal{D}_n) = \int p(y^{\text{test}}|\, x^{\text{test}})\, d\Pi(p|\, x^{\text{test}},\, \mathcal{D}_n)$$

~

PFNs aim at approximating PPDs, since these define distributions that maximize the expectancy of the conditional log-likelihood induced by the prior:

$$\pi(Y^{\text{test}}|\, X^{\text{test}}, \, \mathcal{D}_n) = \underset{q}{\text{argmax}}\, \mathbb{E}_{\Pi}[ \text{log}\, q(Y^{\text{test}}|\, X^{\text{test}}, \, \mathcal{D}_n)]$$

\subsubsection{Approximation of PPDs}

A PFN can be considered as a model $q_{\theta}$ parametrized by a parameter $\theta$. Drawing inspiration from Müller et al. (2021) \cite{muller_transformers_2024}, $\theta$ is optimized by solving:

$$\theta^* = \underset{\theta}{\text{argmax}} \, \mathbb{E}_{\Pi_N} \mathbb{E}_{\Pi} [q_{\theta}(Y^{\text{test}}\, |\, X^{\text{test}}, \, \mathcal{D}_N)]$$

Where $\Pi_N$ is a distribution over the sample size $N$.

~

In practice, a Monte-Carlo integration is performed to approximate $\theta^*$. More explicitly, the following procedure is applied:

~

\begin{enumerate}[label=\Alph*]
    \item For $j \in \{1;2;\dots;J \}$,
    
    \begin{enumerate}[label=\roman*]
        \item Draw a sample size $N_j \sim \Pi_N$;
        \item Draw a model $p \sim \Pi$;
        \item Draw $N_j$ iid samples $\mathcal{D}^{(j)} = (X_{j,i},\, Y_{j,i})_{i=1}^{N_j}$ from model p;
        \item Draw $T_j$ iid test samples $\mathcal{D}^{(j)} = (X_{j,t}^{\text{test}},\, Y_{j,t}^{\text{test}})_{t=1}^{T_j}$ from model p;
    \end{enumerate}

    \item Compute the Negative Log-Likelihood (NLL) loss over the batch of datasets: $$\mathcal{L}(\theta) = - \frac{1}{J} \sum_{j=1}^J \frac{1}{T_j} \sum_{t=1}^{T_j} \text{log} \, q(X_{j,t}^{\text{test}}|\, Y_{j,t}^{\text{test}},\, \mathcal{D}^{(j)})$$
    
    \item Update the parameter $\theta$ via an optimization algorithm like AdamW.
\end{enumerate}

\subsubsection{In-context learning after pretraining}

Once the PFN has been pretrained, no gradient-based optimization is performed anymore on a newly observed dataset. Instead, prediction is carried out entirely through \emph{in-context learning}. In this regime, the calibration samples and the unlabeled test samples are provided together as a single input sequence to the transformer. The model then uses the calibration examples as contextual information from which it infers the predictive relationship to be applied to the test inputs.

~

Formally, let
\[
\mathcal{D}_{\mathrm{train}} = \{(x_i,y_i)\}_{i=1}^{n}
\qquad\text{and}\qquad
\mathcal{D}_{\mathrm{test}} = \{x_j^{\mathrm{test}}\}_{j=1}^{T}.
\]
At inference time, the PFN receives the joint sequence formed by the labeled calibration points and the unlabeled test points. The labels associated with the test inputs are masked and therefore unavailable to the model. The output of the transformer is a predictive distribution for each test sample, conditionally on the whole calibration context:
\[
q_\theta\!\left(
y_1^{\mathrm{test}},\dots,y_T^{\mathrm{test}}
\,\middle|\,
\mathcal{D}_{\mathrm{train}},x_1^{\mathrm{test}},\dots,x_T^{\mathrm{test}}
\right).
\]

In other words, the model does not adapt its weights to the new task; instead, it performs task adaptation internally through the forward pass itself. The calibration set acts as an implicit support set that defines the local prediction problem, while the pretrained transformer applies the statistical regularities learned offline across a large distribution of tasks.

\subsubsection{Transformer view of TabPFN inference}

At a more operational level, TabPFN converts the new dataset into a sequence of tokens processed by a transformer architecture. Each sample is embedded into a latent representation, and labeled calibration samples are represented jointly from their features and targets, whereas test samples are represented from their features only. If $e_x(\cdot)$ and $e_y(\cdot)$ denote feature and label embedding maps, one may schematically write the token representations as
\[
h_i^{\mathrm{train}} = e_x(x_i) + e_y(y_i),
\qquad
h_j^{\mathrm{test}} = e_x(x_j^{\mathrm{test}}) + e_{\mathrm{mask}},
\]
where $e_{\mathrm{mask}}$ is a learned embedding indicating that the target is unknown.

These token representations are then processed through stacked self-attention layers. For a matrix of token states $H \in \mathbb{R}^{m \times p}$, a standard self-attention head computes
\[
Q = HW_Q,\qquad K = HW_K,\qquad V = HW_V,
\]
where $W_Q,W_K,W_V$ are learned projection matrices. The attention output is
\[
\mathrm{Attn}(H)=\mathrm{softmax}\!\left(\frac{QK^\top}{\sqrt{p_k}} + M\right)V,
\]
where $M$ is a masking matrix enforcing the desired information flow. In the PFN setting, this masking mechanism prevents the model from accessing the unknown test labels while allowing test tokens to attend to calibration tokens and, depending on the implementation, to other test inputs.

~

Intuitively, the attention mechanism enables each test token to retrieve relevant information from the labeled calibration examples. Samples with similar feature patterns may exchange more information through larger attention weights, so that the transformer can internally construct a task-adapted predictor from the provided context. Repeating this process over multiple layers allows the network to build increasingly rich interactions between calibration and test points.

~

The final hidden state associated with each test token is then mapped to a predictive distribution. In classification, this typically corresponds to a softmax distribution over classes. In regression-oriented versions of PFNs, the output may instead represent a discretized conditional density over bins, from which moments or point predictions can be derived. This is the mechanism through which TabPFN approximates posterior predictive distributions without retraining on the target dataset.

~

Overall, TabPFN can therefore be viewed as a transformer that has learned, during offline pretraining, how to convert a set of labeled examples into a predictive rule for new unlabeled samples. Its originality does not lie in a dataset-specific optimization step, but in the fact that this optimization has effectively been amortized during pretraining over a large distribution of synthetic supervised tasks.

\clearpage
\subsection{CNN-1D architecture and hyperparameter search space}

The CNN-1D baseline used in this benchmark follows a three-block convolutional architecture 
operating directly on one-dimensional spectral inputs. The model is part of the \texttt{nirs4all} model collection \cite{beurier2025nirs4all}.

~

The architecture consists of the following sequential components. An initial spatial dropout layer 
is applied to the raw input to regularize at the feature level before any convolution. Three 
successive \texttt{Conv1D} blocks then extract spectral features at increasing levels of 
abstraction: the first block uses a large kernel and stride to perform coarse spectral 
summarization, the second applies a wider filter bank for intermediate-scale feature extraction, 
and the third refines the representation with a smaller kernel. Each of the two latter 
convolutional blocks is followed by a normalization layer (either Batch Normalization or Layer 
Normalization). The convolutional output is then flattened and passed through a dense hidden 
layer before a final sigmoid output unit. A standard dropout layer is inserted between the first 
and second convolutional blocks.

~

All architectural components were jointly optimized, 
including filter sizes, kernel sizes, strides, activation functions, normalization methods, 
dropout rates, and the number of dense units.

~

It should be noted that the fixed three-block structure of this architecture implies minimum 
input length constraints that depend on the sampled kernel sizes and strides. For the smallest 
datasets in the benchmark, certain hyperparameter configurations may therefore produce 
architecturally invalid models, in which case the trial is discarded and a new configuration 
is sampled.

\begin{table}[!htbp]
\centering
\caption{Search spaces of hyperparameter configuration of the CNN-1D architecture.}
\label{tab:cnn1d_default}
\small
\renewcommand{\arraystretch}{1.2}
\begin{tabular}{llll}
\toprule
\texttt{Component} & \texttt{Parameter} & \texttt{Search space} & \texttt{Default value} \\
\midrule
Spatial Dropout      & rate          & $\mathcal{U}(0.01,\, 0.50)$         & 0.08 \\
\midrule
Conv1D block 1       & filters       & $\{4, 8, 16, 32, 64, 128, 256\}$        & 8 \\
                     & kernel size   & $\{3, 5, 7, 9, 11, 13, 15\}$        & 15 \\
                     & stride        & $\{1, 2, 3, 4, 5\}$        & 5 \\
                     & activation    & \{ReLU, SELU, ELU, Swish\}        & SELU \\
\midrule
Dropout              & rate          & $\mathcal{U}(0.01,\, 0.50)$        & 0.20 \\
\midrule
Conv1D block 2       & filters       & $\{4, 8, 16, 32, 64, 128, 256\}$        & 64 \\
                     & kernel size   & $\{3, 5, 7, 9, 11, 13, 15\}$        & 21 \\
                     & stride        & $\{1, 2, 3, 4, 5\}$        & 3 \\
                     & activation    & \{ReLU, SELU, ELU, Swish\}        & ReLU \\
\midrule                     
Normalization 1      & method        & \{Batch Norm., Layer Norm.\}        & Batch Normalization \\
\midrule
Conv1D block 3       & filters       & $\{4, 8, 16, 32, 64, 128, 256\}$        & 32 \\
                     & kernel size   & $\{3, 5, 7, 9, 11, 13, 15\}$        & 5 \\
                     & stride        & $\{1, 2, 3, 4, 5\}$        & 3 \\
                     & activation    & \{ReLU, SELU, ELU, Swish\}        & ELU \\
\midrule                     
Normalization 2      & method        & \{Batch Norm., Layer Norm.\}        & Batch Normalization \\
\midrule
Dense hidden layer   & units         & $\{4, 8, 16, 32, 64, 128, 256\}$        & 16 \\
                     & activation    & \{ReLU, SELU, ELU, Swish\}        & Sigmoid \\
\midrule                     
Output layer         & units         & \{1\}        & 1 \\
                     & activation    & \{Sigmoid\}        & Sigmoid \\
                     
\bottomrule
\end{tabular}
\end{table}

\clearpage
\subsection{Additional regression comparison figures}

\begin{figure}[H]
    \centering
    \includegraphics[width=0.92\textwidth,height=0.8\textheight,keepaspectratio]{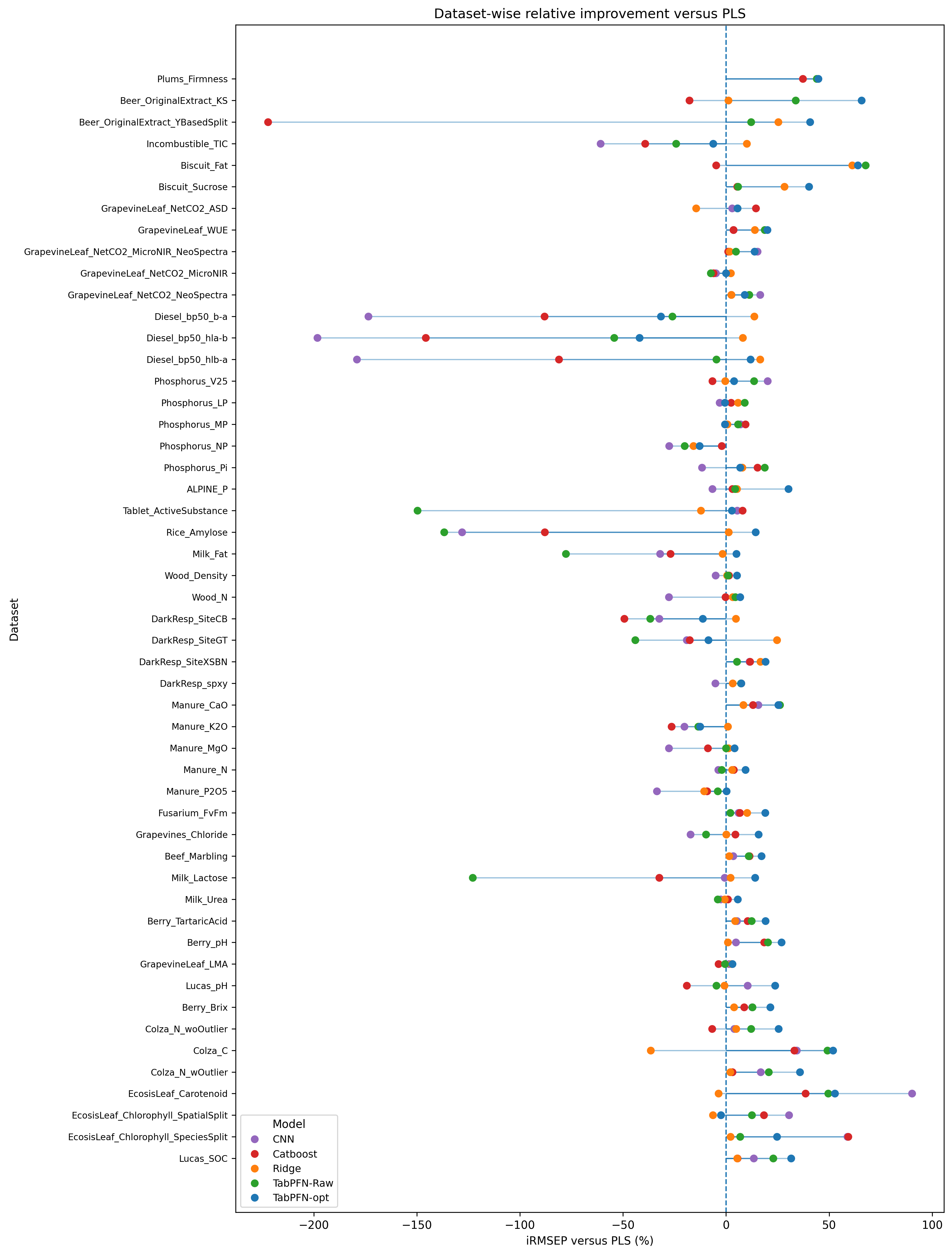}
    \caption{Dataset-wise relative improvement versus PLS for regression tasks, expressed as iRMSEP. Each row corresponds to one dataset and each colored point to one model. Positive values indicate improved performance compared to PLS, whereas negative values indicate worse performance. For visual clarity, the visualization excludes extreme dataset-level iRMSEP values identified by a 10$\times$IQR display filter.}
    \label{fig:dumbbell_regression_irmsep}
\end{figure}

\begin{figure}[p]
\centering
    \includegraphics[width=0.99\textwidth,height=0.78\textheight,keepaspectratio]{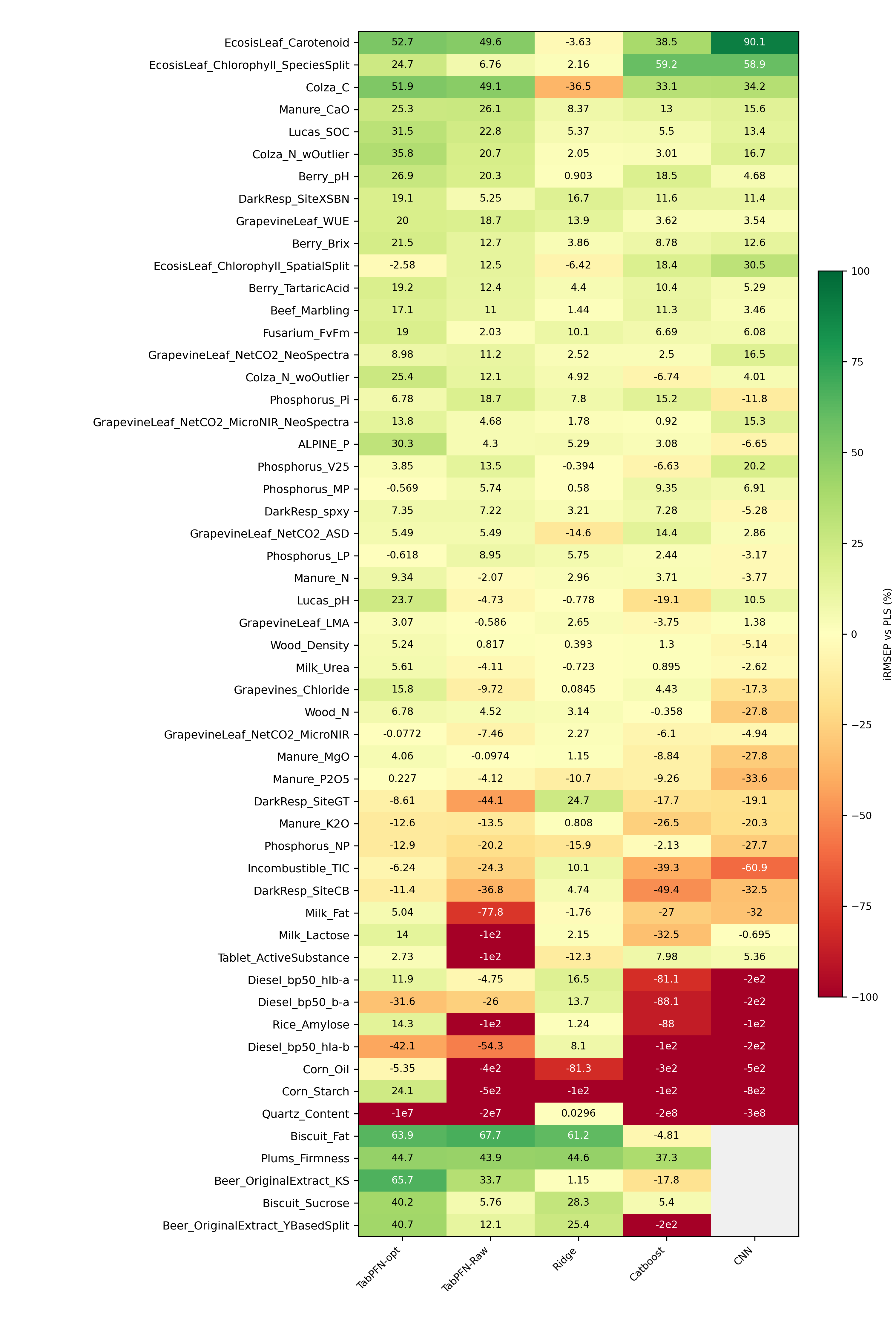}
    \caption{Dataset-level heatmap of iRMSEP relative to PLS for regression tasks. Positive values indicate improved performance compared to PLS, whereas negative values indicate worse performance. Grey cells correspond to unavailable comparisons.}
    \label{fig:heatmap_regression_irmsep_vs_pls}
\end{figure}

\begin{figure}[p]
    \centering
    \includegraphics[width=0.92\textwidth]{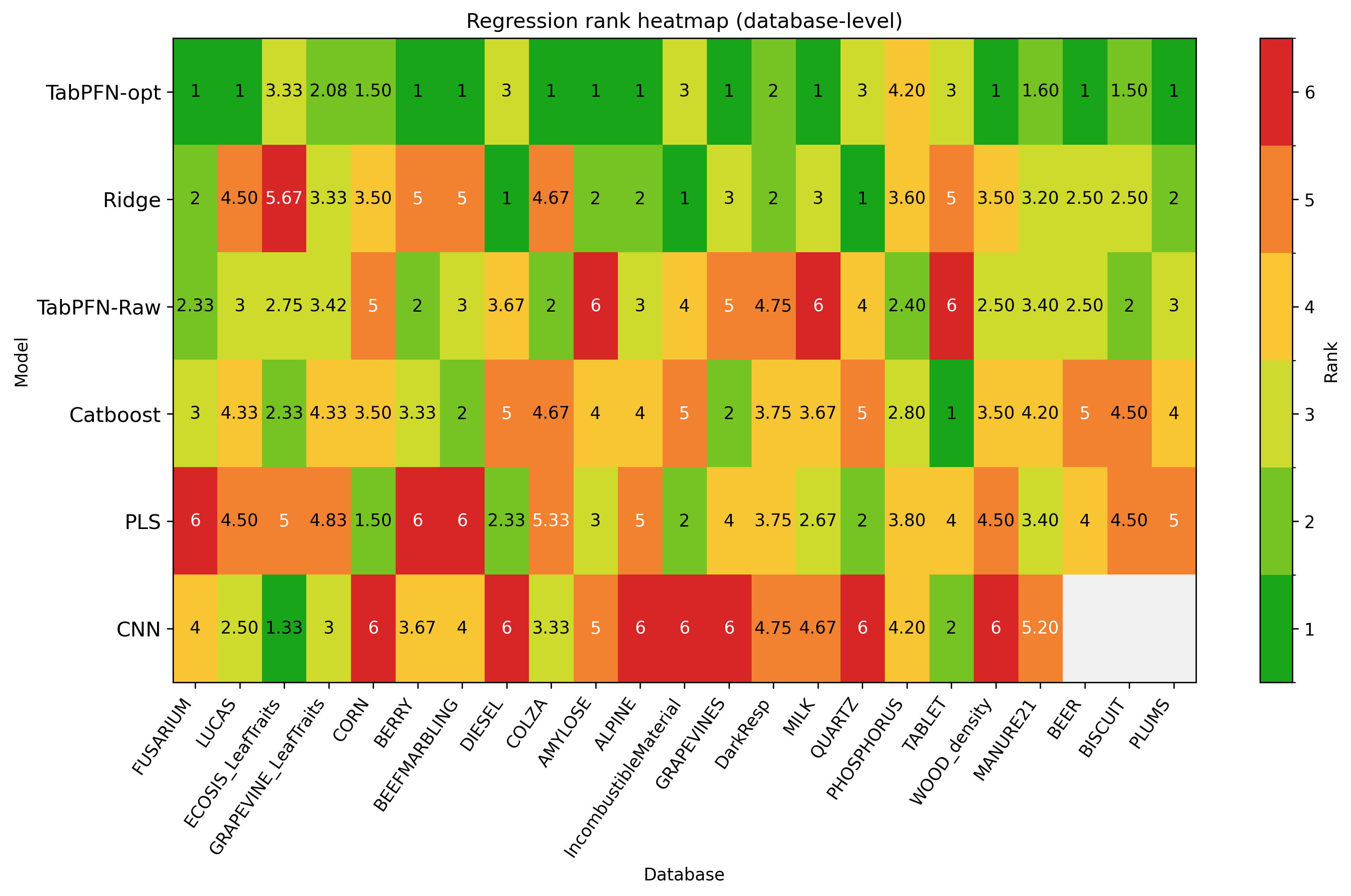}
    \caption{Regression rank heatmap after aggregation at the database level. Lower ranks indicate better predictive performance within each database. Grey cells correspond to unavailable model--database combinations.}
    \label{fig:rank_heatmap_regression_database}
\end{figure}

\begin{figure}[p]
    \centering
    \includegraphics[width=0.92\textwidth,height=0.78\textheight,keepaspectratio]{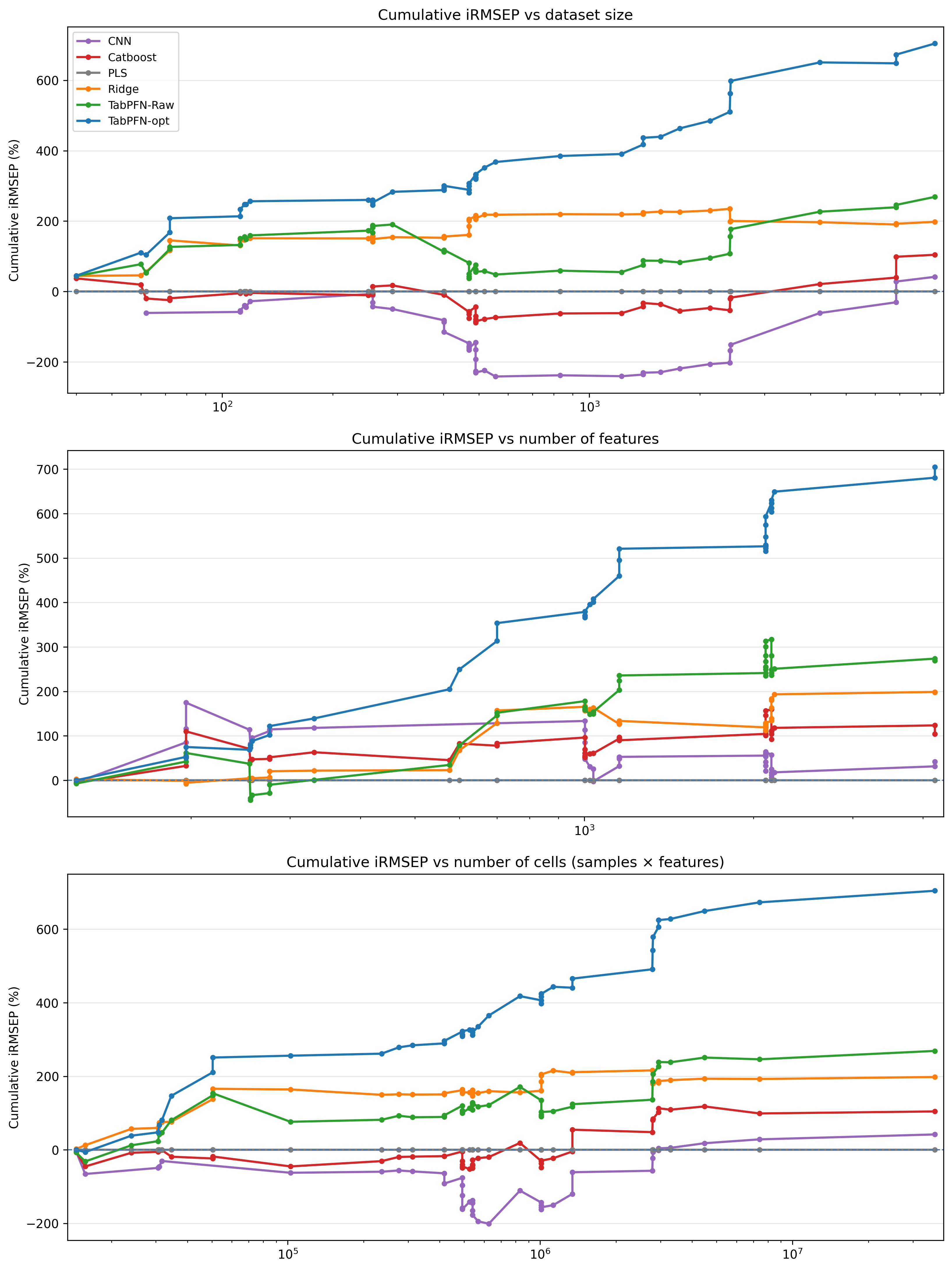}
    \caption{Cumulative iRMSEP values against PLS using 3 different x axis: dataset number of samples, dataset number of features, dataset number of cells (number of samples x number of features). For visual clarity, the visualization excludes extreme dataset-level iRMSEP values identified by a 10$\times$IQR display filter.}
    \label{fig:cumsum_irmsep_size_features_cells}
\end{figure}

\begin{table}[!htbp]
\centering
\caption{Complete statistical summary for regression and classification benchmarks. Mean ranks are reported with and without restricting to datasets where CNN-1D is available. All pairwise Nemenyi comparisons are shown (unique pairs only).}
\label{tab:cd_stats_full}
\scriptsize
\setlength{\tabcolsep}{4pt}
\renewcommand{\arraystretch}{1.05}

\begin{tabular}{lcc}
\toprule
Model & Rank (with CNN-1D) & Rank (w/o CNN-1D) \\
\midrule

\multicolumn{3}{c}{\textbf{Regression}} \\
\midrule
TabPFN-opt & 1.800 & 1.587 \\
Ridge      & 3.075 & 2.696 \\
CatBoost   & 3.650 & 3.587 \\
TabPFN-Raw & 3.875 & 3.370 \\
PLS        & 4.050 & 3.761 \\
CNN-1D     & 4.550 & -- \\

\midrule
\multicolumn{3}{c}{\textbf{Classification}} \\
\midrule
TabPFN-opt & 2.417 & 2.389 \\
TabPFN-Raw & 2.000 & 1.944 \\
CatBoost   & 3.083 & 3.111 \\
PLS-DA     & 2.583 & 2.556 \\
CNN-1D     & 4.917 & -- \\

\bottomrule
\end{tabular}

\vspace{0.35em}

\begin{tabular}{lcc}
\toprule
Model pair & $p$-value & Significance \\
\midrule

\multicolumn{3}{c}{\textbf{Regression}} \\
\midrule
TabPFN-opt vs Ridge      & 0.148 & ns \\
TabPFN-opt vs CatBoost   & 0.0218 & * \\
TabPFN-opt vs TabPFN-Raw & 0.0060 & ** \\
TabPFN-opt vs PLS        & 0.0020 & ** \\
TabPFN-opt vs CNN-1D     & $4.9 \times 10^{-5}$ & **** \\

Ridge vs CatBoost        & 0.707 & ns \\
Ridge vs TabPFN-Raw      & 0.922 & ns \\
Ridge vs PLS             & 0.967 & ns \\
Ridge vs CNN-1D          & 0.375 & ns \\

CatBoost vs TabPFN-Raw   & 0.999 & ns \\
CatBoost vs PLS          & 0.996 & ns \\
CatBoost vs CNN-1D       & 0.777 & ns \\

TabPFN-Raw vs PLS        & 0.999 & ns \\
TabPFN-Raw vs CNN-1D     & 0.650 & ns \\

PLS vs CNN-1D            & 0.533 & ns \\

\midrule
\multicolumn{3}{c}{\textbf{Classification}} \\
\midrule
TabPFN-opt vs TabPFN-Raw & 0.885 & ns \\
TabPFN-opt vs CatBoost   & 0.635 & ns \\
TabPFN-opt vs PLS-DA     & 0.993 & ns \\
TabPFN-opt vs CNN-1D     & 0.049 & * \\

TabPFN-Raw vs CatBoost   & 0.221 & ns \\
TabPFN-Raw vs PLS-DA     & 0.747 & ns \\
TabPFN-Raw vs CNN-1D     & 0.012 & * \\

CatBoost vs PLS-DA       & 0.798 & ns \\
CatBoost vs CNN-1D       & 0.262 & ns \\

PLS-DA vs CNN-1D         & 0.079 & ns \\

\bottomrule
\end{tabular}

\vspace{0.2em}
\begin{minipage}{0.95\linewidth}
\scriptsize
\textit{Notes.} ``With CNN-1D'' corresponds to the strict intersection of datasets where all models provide valid results, while ``without CNN-1D'' includes datasets where CNN-1D may be unavailable. The Friedman test is significant for regression ($p = 5.9 \times 10^{-5}$, $CD = 1.520$) as well as for classification ($p = 0.011$, $CD = 2.345$). Significance codes: ns ($p \geq 0.05$), * ($p < 0.05$), ** ($p < 0.01$), *** ($p < 0.001$), **** ($p < 0.0001$).
\end{minipage}
\end{table}

\FloatBarrier

\clearpage
\subsection{Influence of preprocessings}

\begin{figure}[H]
    \centering
    \includegraphics[width=0.92\textwidth,height=0.8\textheight,keepaspectratio]{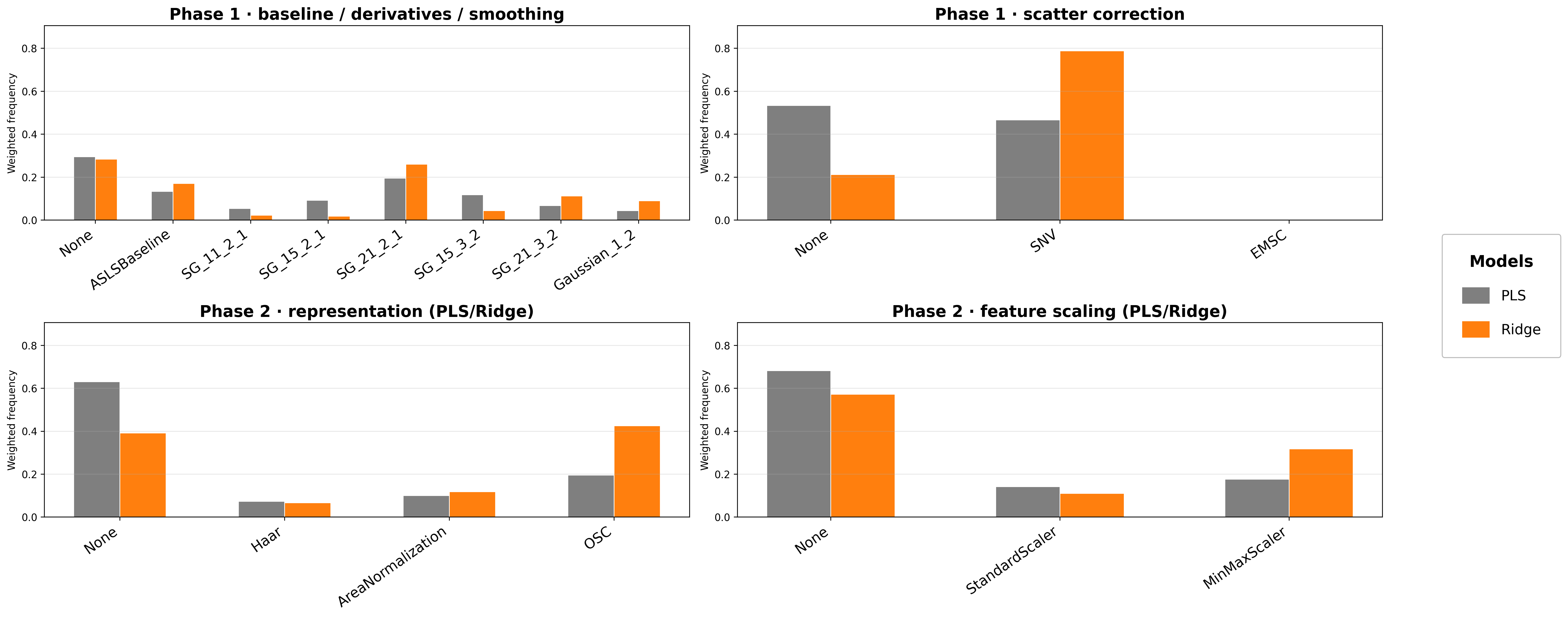}
    \caption{Weighted preprocessing frequencies for linear models (PLS and Ridge) across the regression benchmark. Frequencies are shown separately for Phase~1 signal-correction steps (baseline correction, smoothing, derivatives, scatter correction) and Phase~2 transformations (representation and scaling).}
    \label{fig:preprocessing_linear_models}
\end{figure}

\begin{figure}[p]
    \centering
    \includegraphics[width=0.92\textwidth]{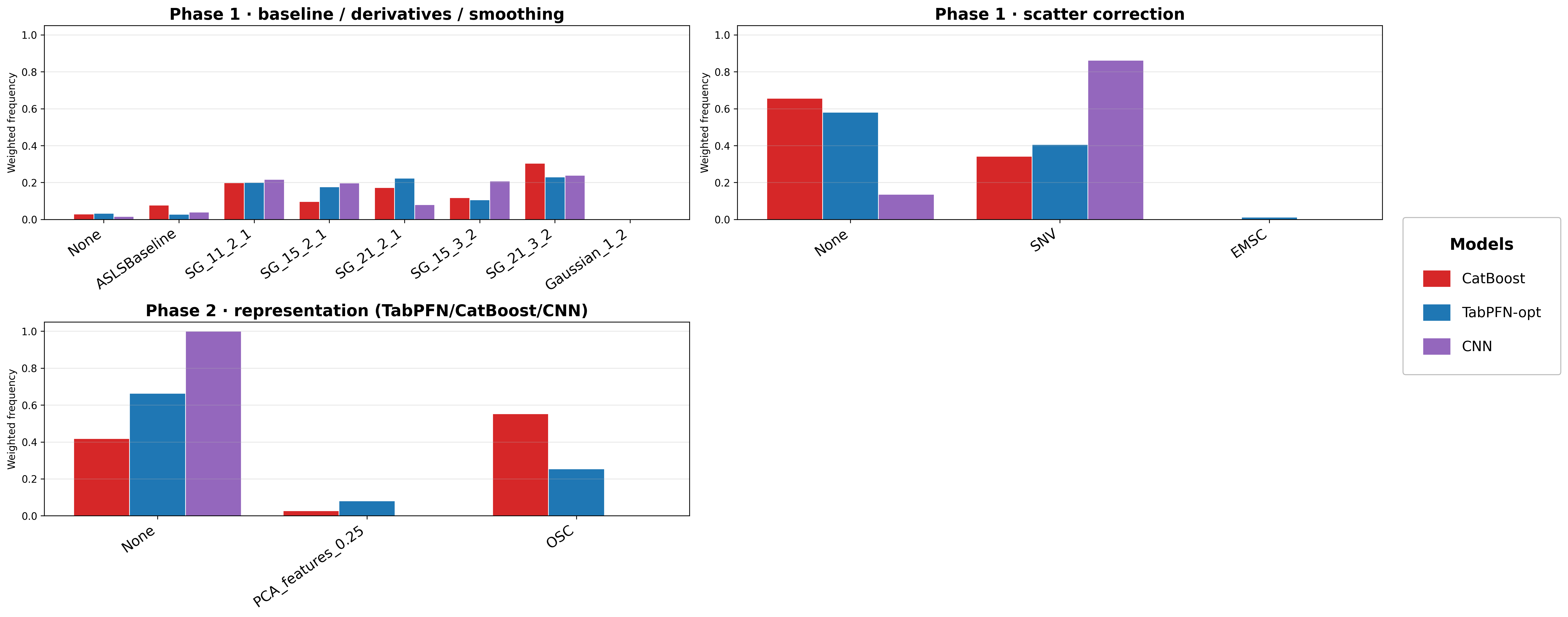}
    \caption{Weighted preprocessing frequencies for nonlinear models (CatBoost, TabPFN, and CNN) across the regression benchmark. Frequencies are shown separately for Phase~1 signal-correction steps and Phase~2 representation-level transformations.}
    \label{fig:preprocessing_tabular_models}
\end{figure}

\FloatBarrier

\clearpage
\subsection{Additional classification comparison figures}

\begin{figure}[H]
    \centering
    \includegraphics[width=0.92\textwidth,height=0.8\textheight,keepaspectratio]{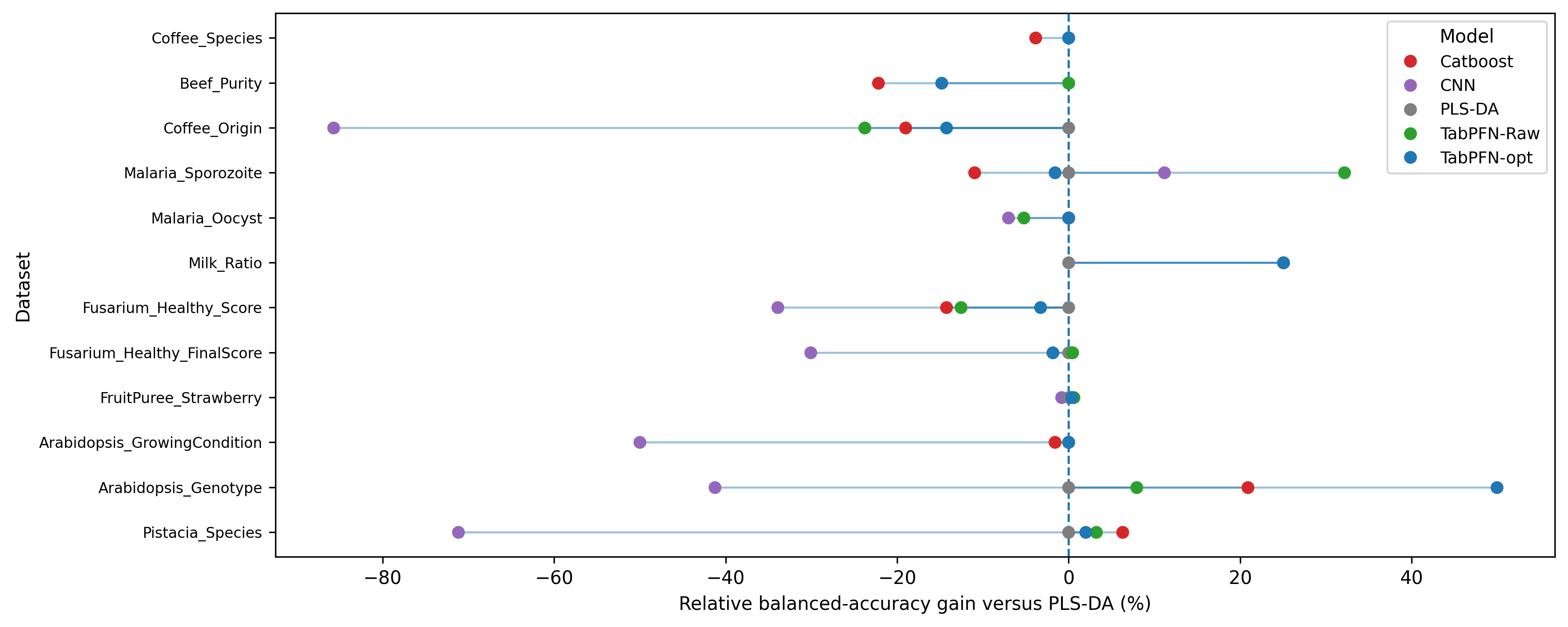}
    \caption{Dataset-wise relative balanced-accuracy gain versus PLS-DA for classification tasks. Each row corresponds to one dataset and each colored point to one model. Positive values indicate improved performance compared to PLS-DA, whereas negative values indicate worse performance. The dashed vertical line indicates parity with PLS-DA.}
    \label{fig:classif_dumbbell}
\end{figure}

\begin{figure}[p]
    \centering
    \includegraphics[width=0.92\textwidth]{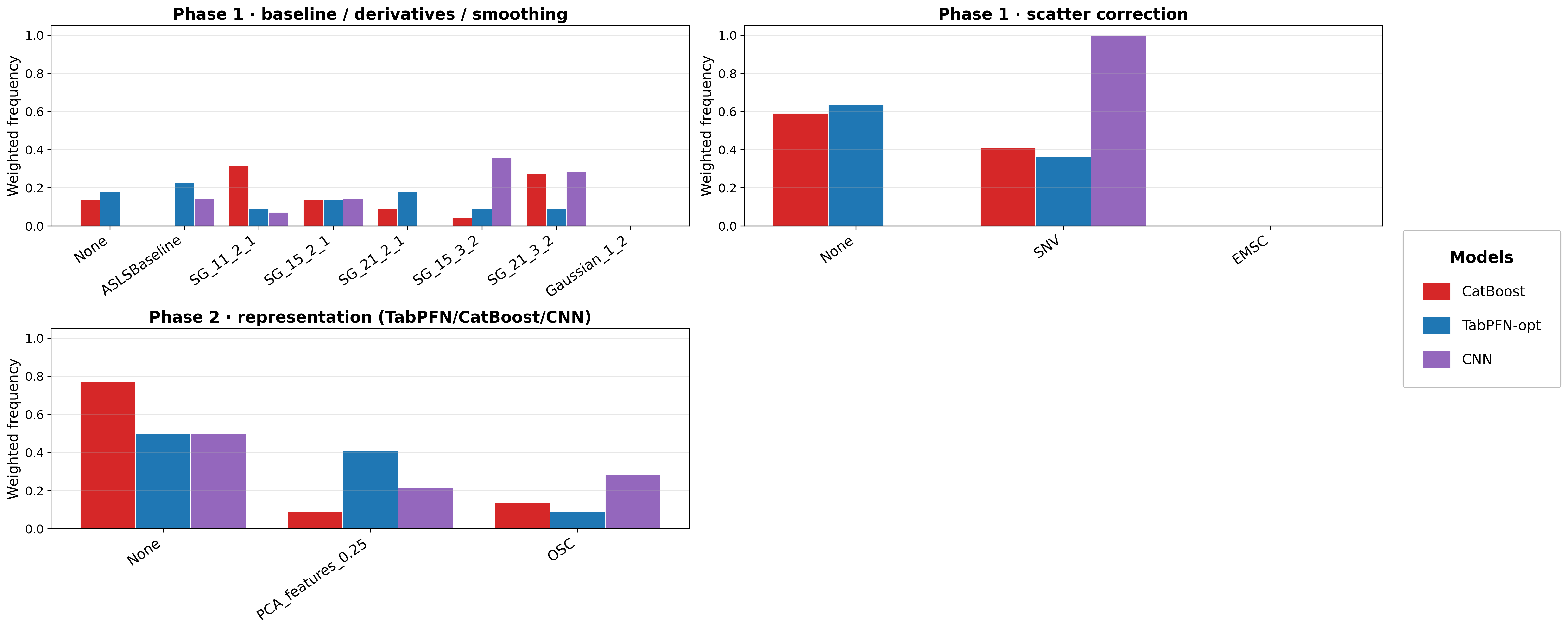}
    \caption{Weighted preprocessing frequencies by search subspace for CatBoost, TabPFN, and CNN in the classification benchmark. Frequencies are shown separately for Phase~1 signal-correction choices and Phase~2 representation-level transformations.}
    \label{fig:classif_preprocessing_tabular_models}
\end{figure}

\FloatBarrier

\clearpage 
\begin{landscape}
\subsection{Complete datasetwise regression results}\label{supp:complete_results_reg}

\begingroup
\setlength{\LTleft}{0pt}
\setlength{\LTright}{0pt}
\begingroup
\small
\renewcommand{\arraystretch}{1.15}
\begin{longtable}{L{4.8cm}L{8.8cm}L{2.3cm}C{1.7cm}C{1.7cm}C{2.0cm}}
\caption{Dataset-wise regression performance for each model. Units are not specified for RMSE values; see the dataset source links for the original units.}\label{tab:complete_results_reg}\\
\toprule
Database & Dataset & Model & RMSECV & RMSEP & iRMSEP (\%) \\
\midrule
\endfirsthead
\toprule
Database & Dataset & Model & RMSECV & RMSEP & iRMSEP (\%) \\
\midrule
\endhead
\multirow{6}{*}{ALPINE} & \multirow{6}{*}{ALPINE\_P} & TabPFN-opt & \textbf{0.074} & \textbf{0.043} & \textbf{30.277} \\
 &  & TabPFN-Raw & --- & 0.060 & 4.300 \\
 &  & Ridge & 0.074 & 0.059 & 5.289 \\
 &  & PLS & 0.077 & 0.062 & 0.000 \\
 &  & CNN & 0.077 & 0.066 & -6.655 \\
 &  & Catboost & 0.080 & 0.060 & 3.080 \\
\addlinespace[0.8em]
\midrule[\heavyrulewidth]
\multirow{6}{*}{AMYLOSE} & \multirow{6}{*}{Rice\_Amylose} & TabPFN-opt & \textbf{2.258} & \textbf{1.632} & \textbf{14.320} \\
 &  & TabPFN-Raw & --- & 4.511 & -136.767 \\
 &  & Ridge & 2.441 & 1.882 & 1.238 \\
 &  & PLS & 2.410 & 1.905 & 0.000 \\
 &  & CNN & 3.743 & 4.345 & -128.073 \\
 &  & Catboost & 3.548 & 3.581 & -87.955 \\
\addlinespace[0.8em]
\midrule[\heavyrulewidth]
\multirow{6}{*}{BEEFMARBLING} & \multirow{6}{*}{Beef\_Marbling} & TabPFN-opt & 63.612 & \textbf{61.661} & \textbf{17.148} \\
 &  & TabPFN-Raw & --- & 66.260 & 10.969 \\
 &  & Ridge & 71.599 & 73.355 & 1.436 \\
 &  & PLS & 71.354 & 74.423 & 0.000 \\
 &  & CNN & 73.179 & 71.850 & 3.458 \\
 &  & Catboost & \textbf{62.891} & 65.979 & 11.347 \\
\addlinespace[0.8em]
\midrule[\heavyrulewidth]
\multirow{10}{*}{BEER} & \multirow{5}{*}{Beer\_OriginalExtract\_KS} & TabPFN-opt & \textbf{0.332} & \textbf{0.130} & \textbf{65.682} \\
 &  & TabPFN-Raw & --- & 0.251 & 33.695 \\
 &  & Ridge & 0.442 & 0.374 & 1.151 \\
 &  & PLS & 0.442 & 0.379 & 0.000 \\
 &  & Catboost & 1.298 & 0.446 & -17.762 \\
\addlinespace[0.8em]
 & \multirow{5}{*}{Beer\_OriginalExtract\_YBasedSplit} & TabPFN-opt & 0.414 & \textbf{0.152} & \textbf{40.686} \\
 &  & TabPFN-Raw & --- & 0.225 & 12.141 \\
 &  & Ridge & \textbf{0.278} & 0.191 & 25.370 \\
 &  & PLS & 0.373 & 0.256 & 0.000 \\
 &  & Catboost & 1.285 & 0.825 & -222.301 \\
\addlinespace[0.8em]
\midrule[\heavyrulewidth]
\multirow{18}{*}{BERRY} & \multirow{6}{*}{Berry\_Brix} & TabPFN-opt & \textbf{2.860} & \textbf{2.853} & \textbf{21.498} \\
 &  & TabPFN-Raw & --- & 3.171 & 12.749 \\
 &  & Ridge & 3.563 & 3.494 & 3.863 \\
 &  & PLS & 3.803 & 3.635 & 0.000 \\
 &  & CNN & 3.477 & 3.178 & 12.567 \\
 &  & Catboost & 3.267 & 3.315 & 8.779 \\
\addlinespace[0.8em]
 & \multirow{6}{*}{Berry\_pH} & TabPFN-opt & \textbf{0.213} & \textbf{0.231} & \textbf{26.900} \\
 &  & TabPFN-Raw & --- & 0.252 & 20.310 \\
 &  & Ridge & 0.271 & 0.313 & 0.903 \\
 &  & PLS & 0.287 & 0.316 & 0.000 \\
 &  & CNN & 0.266 & 0.301 & 4.677 \\
 &  & Catboost & 0.241 & 0.258 & 18.456 \\
\addlinespace[0.8em]
 & \multirow{6}{*}{Berry\_TartaricAcid} & TabPFN-opt & \textbf{1.545} & \textbf{1.523} & \textbf{19.159} \\
 &  & TabPFN-Raw & --- & 1.651 & 12.394 \\
 &  & Ridge & 1.716 & 1.801 & 4.404 \\
 &  & PLS & 1.788 & 1.884 & 0.000 \\
 &  & CNN & 1.691 & 1.785 & 5.289 \\
 &  & Catboost & 1.633 & 1.688 & 10.399 \\
\addlinespace[0.8em]
\midrule[\heavyrulewidth]
\multirow{10}{*}{BISCUIT} & \multirow{5}{*}{Biscuit\_Fat} & TabPFN-opt & 0.405 & 0.450 & 63.914 \\
 &  & TabPFN-Raw & --- & \textbf{0.403} & \textbf{67.694} \\
 &  & Ridge & 0.393 & 0.483 & 61.204 \\
 &  & PLS & \textbf{0.352} & 1.246 & 0.000 \\
 &  & Catboost & 0.892 & 1.306 & -4.806 \\
\addlinespace[0.8em]
 & \multirow{5}{*}{Biscuit\_Sucrose} & TabPFN-opt & \textbf{1.671} & \textbf{0.956} & \textbf{40.179} \\
 &  & TabPFN-Raw & --- & 1.507 & 5.758 \\
 &  & Ridge & 1.925 & 1.146 & 28.303 \\
 &  & PLS & 1.857 & 1.599 & 0.000 \\
 &  & Catboost & 1.926 & 1.512 & 5.396 \\
\addlinespace[0.8em]
\midrule[\heavyrulewidth]
\multirow{18}{*}{COLZA} & \multirow{6}{*}{Colza\_C} & TabPFN-opt & \textbf{0.960} & \textbf{0.793} & \textbf{51.856} \\
 &  & TabPFN-Raw & --- & 0.839 & 49.114 \\
 &  & Ridge & 1.283 & 2.250 & -36.523 \\
 &  & PLS & 1.338 & 1.648 & 0.000 \\
 &  & CNN & 1.269 & 1.084 & 34.244 \\
 &  & Catboost & 1.137 & 1.103 & 33.080 \\
\addlinespace[0.8em]
 & \multirow{6}{*}{Colza\_N\_wOutlier} & TabPFN-opt & \textbf{0.198} & \textbf{0.175} & \textbf{35.850} \\
 &  & TabPFN-Raw & --- & 0.217 & 20.712 \\
 &  & Ridge & 0.336 & 0.268 & 2.051 \\
 &  & PLS & 0.338 & 0.273 & 0.000 \\
 &  & CNN & 0.287 & 0.228 & 16.742 \\
 &  & Catboost & 0.275 & 0.265 & 3.014 \\
\addlinespace[0.8em]
 & \multirow{6}{*}{Colza\_N\_woOutlier} & TabPFN-opt & \textbf{0.181} & \textbf{0.177} & \textbf{25.449} \\
 &  & TabPFN-Raw & --- & 0.209 & 12.101 \\
 &  & Ridge & 0.235 & 0.226 & 4.918 \\
 &  & PLS & 0.246 & 0.238 & 0.000 \\
 &  & CNN & 0.257 & 0.228 & 4.009 \\
 &  & Catboost & 0.252 & 0.254 & -6.740 \\
\addlinespace[0.8em]
\midrule[\heavyrulewidth]
\multirow{12}{*}{CORN} & \multirow{6}{*}{Corn\_Oil} & TabPFN-opt & 0.037 & 0.025 & -5.349 \\
 &  & TabPFN-Raw & --- & 0.110 & -369.530 \\
 &  & Ridge & 0.065 & 0.043 & -81.253 \\
 &  & PLS & \textbf{0.026} & \textbf{0.023} & \textbf{0.000} \\
 &  & CNN & 0.118 & 0.140 & -496.540 \\
 &  & Catboost & 0.094 & 0.093 & -298.251 \\
\addlinespace[0.8em]
 & \multirow{6}{*}{Corn\_Starch} & TabPFN-opt & 0.136 & \textbf{0.064} & \textbf{24.143} \\
 &  & TabPFN-Raw & --- & 0.497 & -489.354 \\
 &  & Ridge & 0.196 & 0.197 & -133.291 \\
 &  & PLS & \textbf{0.100} & 0.084 & 0.000 \\
 &  & CNN & 0.742 & 0.739 & -775.701 \\
 &  & Catboost & 0.365 & 0.190 & -125.515 \\
\addlinespace[0.8em]
\midrule[\heavyrulewidth]
\multirow{18}{*}{DIESEL} & \multirow{6}{*}{Diesel\_bp50\_b-a} & TabPFN-opt & 4.535 & 4.328 & -31.630 \\
 &  & TabPFN-Raw & --- & 4.144 & -26.047 \\
 &  & Ridge & 2.759 & \textbf{2.838} & \textbf{13.671} \\
 &  & PLS & \textbf{2.497} & 3.288 & 0.000 \\
 &  & CNN & 12.779 & 8.996 & -173.608 \\
 &  & Catboost & 7.726 & 6.186 & -88.150 \\
\addlinespace[0.8em]
 & \multirow{6}{*}{Diesel\_bp50\_hla-b} & TabPFN-opt & 4.761 & 4.205 & -42.062 \\
 &  & TabPFN-Raw & --- & 4.568 & -54.342 \\
 &  & Ridge & 3.408 & \textbf{2.720} & \textbf{8.101} \\
 &  & PLS & \textbf{3.339} & 2.960 & 0.000 \\
 &  & CNN & 9.226 & 8.831 & -198.364 \\
 &  & Catboost & 7.322 & 7.274 & -145.776 \\
\addlinespace[0.8em]
 & \multirow{6}{*}{Diesel\_bp50\_hlb-a} & TabPFN-opt & 4.802 & 2.942 & 11.850 \\
 &  & TabPFN-Raw & --- & 3.496 & -4.746 \\
 &  & Ridge & 2.985 & \textbf{2.788} & \textbf{16.456} \\
 &  & PLS & \textbf{2.900} & 3.337 & 0.000 \\
 &  & CNN & 11.197 & 9.316 & -179.163 \\
 &  & Catboost & 7.421 & 6.045 & -81.150 \\
\addlinespace[0.8em]
\midrule[\heavyrulewidth]
\multirow{24}{*}{DarkResp} & \multirow{6}{*}{DarkResp\_SiteCB} & TabPFN-opt & 0.183 & 0.277 & -11.354 \\
 &  & TabPFN-Raw & --- & 0.340 & -36.792 \\
 &  & Ridge & \textbf{0.177} & \textbf{0.237} & \textbf{4.738} \\
 &  & PLS & 0.178 & 0.249 & 0.000 \\
 &  & CNN & 0.200 & 0.329 & -32.483 \\
 &  & Catboost & 0.184 & 0.371 & -49.433 \\
\addlinespace[0.8em]
 & \multirow{6}{*}{DarkResp\_SiteGT} & TabPFN-opt & \textbf{0.194} & 0.284 & -8.611 \\
 &  & TabPFN-Raw & --- & 0.376 & -44.076 \\
 &  & Ridge & 0.199 & \textbf{0.197} & \textbf{24.651} \\
 &  & PLS & 0.200 & 0.261 & 0.000 \\
 &  & CNN & 0.219 & 0.311 & -19.095 \\
 &  & Catboost & 0.202 & 0.307 & -17.692 \\
\addlinespace[0.8em]
 & \multirow{6}{*}{DarkResp\_SiteXSBN} & TabPFN-opt & 0.186 & \textbf{0.256} & \textbf{19.076} \\
 &  & TabPFN-Raw & --- & 0.299 & 5.250 \\
 &  & Ridge & 0.193 & 0.263 & 16.684 \\
 &  & PLS & 0.194 & 0.316 & 0.000 \\
 &  & CNN & 0.206 & 0.280 & 11.394 \\
 &  & Catboost & \textbf{0.185} & 0.279 & 11.555 \\
\addlinespace[0.8em]
 & \multirow{6}{*}{DarkResp\_spxy} & TabPFN-opt & 0.200 & \textbf{0.166} & \textbf{7.345} \\
 &  & TabPFN-Raw & --- & 0.166 & 7.222 \\
 &  & Ridge & \textbf{0.197} & 0.174 & 3.213 \\
 &  & PLS & 0.197 & 0.179 & 0.000 \\
 &  & CNN & 0.226 & 0.189 & -5.279 \\
 &  & Catboost & 0.213 & 0.166 & 7.279 \\
\addlinespace[0.8em]
\midrule[\heavyrulewidth]
\multirow{18}{*}{ECOSIS\_LeafTraits} & \multirow{6}{*}{EcosisLeaf\_Carotenoid} & TabPFN-opt & 4.654 & 30.400 & 52.717 \\
 &  & TabPFN-Raw & --- & 32.424 & 49.570 \\
 &  & Ridge & 35.712 & 66.631 & -3.635 \\
 &  & PLS & 36.719 & 64.294 & 0.000 \\
 &  & CNN & \textbf{3.926} & \textbf{6.338} & \textbf{90.141} \\
 &  & Catboost & 16.276 & 39.548 & 38.489 \\
\addlinespace[0.8em]
 & \multirow{6}{*}{EcosisLeaf\_Chlorophyll\_SpatialSplit} & TabPFN-opt & \textbf{7.844} & 70.252 & -2.583 \\
 &  & TabPFN-Raw & --- & 59.898 & 12.536 \\
 &  & Ridge & 19.001 & 72.881 & -6.422 \\
 &  & PLS & 19.643 & 68.483 & 0.000 \\
 &  & CNN & 10.347 & \textbf{47.589} & \textbf{30.510} \\
 &  & Catboost & 11.774 & 55.893 & 18.385 \\
\addlinespace[0.8em]
 & \multirow{6}{*}{EcosisLeaf\_Chlorophyll\_SpeciesSplit} & TabPFN-opt & \textbf{6.999} & 49.003 & 24.730 \\
 &  & TabPFN-Raw & --- & 60.699 & 6.764 \\
 &  & Ridge & 17.204 & 63.699 & 2.156 \\
 &  & PLS & 17.974 & 65.102 & 0.000 \\
 &  & CNN & 10.703 & 26.785 & 58.858 \\
 &  & Catboost & 10.036 & \textbf{26.550} & \textbf{59.219} \\
\addlinespace[0.8em]
\midrule[\heavyrulewidth]
\multirow{6}{*}{FUSARIUM} & \multirow{6}{*}{Fusarium\_FvFm} & TabPFN-opt & \textbf{0.035} & \textbf{0.026} & \textbf{19.037} \\
 &  & TabPFN-Raw & --- & 0.031 & 2.027 \\
 &  & Ridge & 0.036 & 0.028 & 10.122 \\
 &  & PLS & 0.037 & 0.032 & 0.000 \\
 &  & CNN & 0.037 & 0.030 & 6.084 \\
 &  & Catboost & 0.035 & 0.029 & 6.692 \\
\addlinespace[0.8em]
\midrule[\heavyrulewidth]
\multirow{6}{*}{GRAPEVINES} & \multirow{6}{*}{Grapevines\_Chloride} & TabPFN-opt & \textbf{899.616} & \textbf{771.275} & \textbf{15.775} \\
 &  & TabPFN-Raw & --- & 1004.771 & -9.723 \\
 &  & Ridge & 995.711 & 914.961 & 0.085 \\
 &  & PLS & 1025.010 & 915.735 & 0.000 \\
 &  & CNN & 1108.102 & 1073.720 & -17.252 \\
 &  & Catboost & 967.768 & 875.196 & 4.427 \\
\addlinespace[0.8em]
\midrule[\heavyrulewidth]
\multirow{36}{*}{GRAPEVINE\_LeafTraits} & \multirow{6}{*}{GrapevineLeaf\_NetCO2\_ASD} & TabPFN-opt & 3.037 & 3.420 & 5.489 \\
 &  & TabPFN-Raw & --- & 3.420 & 5.489 \\
 &  & Ridge & \textbf{2.929} & 4.146 & -14.596 \\
 &  & PLS & 2.999 & 3.618 & 0.000 \\
 &  & CNN & 3.228 & 3.515 & 2.863 \\
 &  & Catboost & 3.100 & \textbf{3.097} & \textbf{14.397} \\
\addlinespace[0.8em]
 & \multirow{6}{*}{GrapevineLeaf\_NetCO2\_MicroNIR} & TabPFN-opt & 3.273 & 3.735 & -0.077 \\
 &  & TabPFN-Raw & --- & 4.010 & -7.459 \\
 &  & Ridge & \textbf{3.200} & \textbf{3.647} & \textbf{2.274} \\
 &  & PLS & 3.254 & 3.732 & 0.000 \\
 &  & CNN & 3.469 & 3.916 & -4.937 \\
 &  & Catboost & 3.292 & 3.960 & -6.100 \\
\addlinespace[0.8em]
 & \multirow{6}{*}{GrapevineLeaf\_NetCO2\_MicroNIR\_NeoSpectra} & TabPFN-opt & 3.444 & 3.624 & 13.764 \\
 &  & TabPFN-Raw & --- & 4.005 & 4.684 \\
 &  & Ridge & \textbf{3.156} & 4.127 & 1.783 \\
 &  & PLS & 3.181 & 4.202 & 0.000 \\
 &  & CNN & 3.645 & \textbf{3.560} & \textbf{15.282} \\
 &  & Catboost & 3.446 & 4.163 & 0.920 \\
\addlinespace[0.8em]
 & \multirow{6}{*}{GrapevineLeaf\_NetCO2\_NeoSpectra} & TabPFN-opt & 3.333 & 4.485 & 8.984 \\
 &  & TabPFN-Raw & --- & 4.375 & 11.216 \\
 &  & Ridge & \textbf{3.233} & 4.803 & 2.523 \\
 &  & PLS & 3.408 & 4.927 & 0.000 \\
 &  & CNN & 3.352 & \textbf{4.115} & \textbf{16.480} \\
 &  & Catboost & 3.331 & 4.804 & 2.497 \\
\addlinespace[0.8em]
 & \multirow{6}{*}{GrapevineLeaf\_LMA} & TabPFN-opt & 0.293 & \textbf{0.303} & \textbf{3.069} \\
 &  & TabPFN-Raw & --- & 0.314 & -0.586 \\
 &  & Ridge & \textbf{0.282} & 0.304 & 2.653 \\
 &  & PLS & 0.286 & 0.312 & 0.000 \\
 &  & CNN & 0.292 & 0.308 & 1.377 \\
 &  & Catboost & 0.309 & 0.324 & -3.750 \\
\addlinespace[0.8em]
 & \multirow{6}{*}{GrapevineLeaf\_WUE} & TabPFN-opt & 1.674 & \textbf{1.483} & \textbf{19.973} \\
 &  & TabPFN-Raw & --- & 1.507 & 18.683 \\
 &  & Ridge & 1.665 & 1.595 & 13.936 \\
 &  & PLS & \textbf{1.650} & 1.853 & 0.000 \\
 &  & CNN & 1.971 & 1.788 & 3.536 \\
 &  & Catboost & 1.795 & 1.786 & 3.617 \\
\addlinespace[0.8em]
\midrule[\heavyrulewidth]
\multirow{6}{*}{IncombustibleMaterial} & \multirow{6}{*}{Incombustible\_TIC} & TabPFN-opt & 5.379 & 2.962 & -6.243 \\
 &  & TabPFN-Raw & --- & 3.464 & -24.259 \\
 &  & Ridge & \textbf{4.348} & \textbf{2.507} & \textbf{10.077} \\
 &  & PLS & 5.381 & 2.788 & 0.000 \\
 &  & CNN & 6.828 & 4.485 & -60.871 \\
 &  & Catboost & 5.370 & 3.884 & -39.309 \\
\addlinespace[0.8em]
\midrule[\heavyrulewidth]
\multirow{12}{*}{LUCAS} & \multirow{6}{*}{Lucas\_SOC} & TabPFN-opt & \textbf{7.129} & \textbf{3.331} & \textbf{31.484} \\
 &  & TabPFN-Raw & --- & 3.752 & 22.809 \\
 &  & Ridge & 9.708 & 4.600 & 5.373 \\
 &  & PLS & 9.979 & 4.861 & 0.000 \\
 &  & CNN & 8.736 & 4.209 & 13.415 \\
 &  & Catboost & 8.825 & 4.594 & 5.500 \\
\addlinespace[0.8em]
 & \multirow{6}{*}{Lucas\_pH} & TabPFN-opt & 0.349 & \textbf{0.250} & \textbf{23.730} \\
 &  & TabPFN-Raw & --- & 0.344 & -4.728 \\
 &  & Ridge & \textbf{0.345} & 0.331 & -0.778 \\
 &  & PLS & 0.356 & 0.328 & 0.000 \\
 &  & CNN & 0.374 & 0.294 & 10.487 \\
 &  & Catboost & 0.445 & 0.391 & -19.096 \\
\addlinespace[0.8em]
\midrule[\heavyrulewidth]
\multirow{30}{*}{MANURE21} & \multirow{6}{*}{Manure\_CaO} & TabPFN-opt & \textbf{9.504} & 5.622 & 25.328 \\
 &  & TabPFN-Raw & --- & \textbf{5.562} & \textbf{26.125} \\
 &  & Ridge & 11.434 & 6.898 & 8.370 \\
 &  & PLS & 11.672 & 7.529 & 0.000 \\
 &  & CNN & 11.742 & 6.357 & 15.558 \\
 &  & Catboost & 12.193 & 6.549 & 13.017 \\
\addlinespace[0.8em]
 & \multirow{6}{*}{Manure\_K2O} & TabPFN-opt & 3.523 & 2.658 & -12.598 \\
 &  & TabPFN-Raw & --- & 2.679 & -13.494 \\
 &  & Ridge & 3.496 & \textbf{2.342} & \textbf{0.808} \\
 &  & PLS & \textbf{3.478} & 2.361 & 0.000 \\
 &  & CNN & 4.085 & 2.840 & -20.292 \\
 &  & Catboost & 3.789 & 2.987 & -26.504 \\
\addlinespace[0.8em]
 & \multirow{6}{*}{Manure\_MgO} & TabPFN-opt & \textbf{1.059} & \textbf{0.751} & \textbf{4.061} \\
 &  & TabPFN-Raw & --- & 0.783 & -0.097 \\
 &  & Ridge & 1.089 & 0.774 & 1.151 \\
 &  & PLS & 1.095 & 0.783 & 0.000 \\
 &  & CNN & 1.432 & 1.001 & -27.833 \\
 &  & Catboost & 1.241 & 0.852 & -8.843 \\
\addlinespace[0.8em]
 & \multirow{6}{*}{Manure\_P2O5} & TabPFN-opt & \textbf{3.294} & \textbf{2.295} & \textbf{0.227} \\
 &  & TabPFN-Raw & --- & 2.395 & -4.117 \\
 &  & Ridge & 3.426 & 2.547 & -10.718 \\
 &  & PLS & 3.456 & 2.300 & 0.000 \\
 &  & CNN & 3.635 & 3.074 & -33.646 \\
 &  & Catboost & 3.779 & 2.513 & -9.263 \\
\addlinespace[0.8em]
 & \multirow{6}{*}{Manure\_N} & TabPFN-opt & \textbf{2.176} & \textbf{1.594} & \textbf{9.344} \\
 &  & TabPFN-Raw & --- & 1.795 & -2.075 \\
 &  & Ridge & 2.204 & 1.706 & 2.959 \\
 &  & PLS & 2.211 & 1.758 & 0.000 \\
 &  & CNN & 2.322 & 1.825 & -3.770 \\
 &  & Catboost & 2.326 & 1.693 & 3.709 \\
\addlinespace[0.8em]
\midrule[\heavyrulewidth]
\multirow{18}{*}{MILK} & \multirow{6}{*}{Milk\_Fat} & TabPFN-opt & 0.078 & \textbf{0.080} & \textbf{5.041} \\
 &  & TabPFN-Raw & --- & 0.151 & -77.780 \\
 &  & Ridge & \textbf{0.065} & 0.086 & -1.763 \\
 &  & PLS & 0.065 & 0.085 & 0.000 \\
 &  & CNN & 0.207 & 0.112 & -32.026 \\
 &  & Catboost & 0.148 & 0.108 & -26.964 \\
\addlinespace[0.8em]
 & \multirow{6}{*}{Milk\_Lactose} & TabPFN-opt & \textbf{0.055} & \textbf{0.048} & \textbf{14.025} \\
 &  & TabPFN-Raw & --- & 0.124 & -122.887 \\
 &  & Ridge & 0.060 & 0.054 & 2.146 \\
 &  & PLS & 0.059 & 0.055 & 0.000 \\
 &  & CNN & 0.064 & 0.056 & -0.695 \\
 &  & Catboost & 0.087 & 0.073 & -32.467 \\
\addlinespace[0.8em]
 & \multirow{6}{*}{Milk\_Urea} & TabPFN-opt & \textbf{4.115} & \textbf{3.928} & \textbf{5.608} \\
 &  & TabPFN-Raw & --- & 4.332 & -4.113 \\
 &  & Ridge & 4.294 & 4.191 & -0.723 \\
 &  & PLS & 4.251 & 4.161 & 0.000 \\
 &  & CNN & 4.239 & 4.270 & -2.621 \\
 &  & Catboost & 4.492 & 4.124 & 0.895 \\
\addlinespace[0.8em]
\midrule[\heavyrulewidth]
\multirow{30}{*}{PHOSPHORUS} & \multirow{6}{*}{Phosphorus\_LP} & TabPFN-opt & \textbf{0.039} & 0.183 & -0.618 \\
 &  & TabPFN-Raw & --- & \textbf{0.166} & \textbf{8.947} \\
 &  & Ridge & 0.052 & 0.172 & 5.747 \\
 &  & PLS & 0.050 & 0.182 & 0.000 \\
 &  & CNN & 0.053 & 0.188 & -3.168 \\
 &  & Catboost & 0.046 & 0.178 & 2.436 \\
\addlinespace[0.8em]
 & \multirow{6}{*}{Phosphorus\_MP} & TabPFN-opt & \textbf{0.011} & 0.021 & -0.569 \\
 &  & TabPFN-Raw & --- & 0.020 & 5.739 \\
 &  & Ridge & 0.011 & 0.021 & 0.580 \\
 &  & PLS & 0.011 & 0.021 & 0.000 \\
 &  & CNN & 0.011 & 0.020 & 6.906 \\
 &  & Catboost & 0.011 & \textbf{0.019} & \textbf{9.346} \\
\addlinespace[0.8em]
 & \multirow{6}{*}{Phosphorus\_NP} & TabPFN-opt & \textbf{0.034} & 0.116 & -12.940 \\
 &  & TabPFN-Raw & --- & 0.123 & -20.176 \\
 &  & Ridge & 0.046 & 0.119 & -15.855 \\
 &  & PLS & 0.045 & \textbf{0.102} & \textbf{0.000} \\
 &  & CNN & 0.044 & 0.131 & -27.679 \\
 &  & Catboost & 0.040 & 0.105 & -2.127 \\
\addlinespace[0.8em]
 & \multirow{6}{*}{Phosphorus\_Pi} & TabPFN-opt & \textbf{0.036} & 0.180 & 6.782 \\
 &  & TabPFN-Raw & --- & \textbf{0.157} & \textbf{18.745} \\
 &  & Ridge & 0.042 & 0.178 & 7.797 \\
 &  & PLS & 0.042 & 0.193 & 0.000 \\
 &  & CNN & 0.040 & 0.216 & -11.755 \\
 &  & Catboost & 0.040 & 0.164 & 15.163 \\
\addlinespace[0.8em]
 & \multirow{6}{*}{Phosphorus\_V25} & TabPFN-opt & 0.157 & 0.293 & 3.852 \\
 &  & TabPFN-Raw & --- & 0.263 & 13.525 \\
 &  & Ridge & \textbf{0.152} & 0.306 & -0.394 \\
 &  & PLS & 0.154 & 0.304 & 0.000 \\
 &  & CNN & 0.169 & \textbf{0.243} & \textbf{20.193} \\
 &  & Catboost & 0.168 & 0.325 & -6.630 \\
\addlinespace[0.8em]
\midrule[\heavyrulewidth]
\multirow{5}{*}{PLUMS} & \multirow{5}{*}{Plums\_Firmness} & TabPFN-opt & 0.358 & \textbf{0.257} & \textbf{44.711} \\
 &  & TabPFN-Raw & --- & 0.261 & 43.910 \\
 &  & Ridge & 0.368 & 0.258 & 44.604 \\
 &  & PLS & \textbf{0.344} & 0.465 & 0.000 \\
 &  & Catboost & 0.345 & 0.292 & 37.286 \\
\addlinespace[0.8em]
\midrule[\heavyrulewidth]
\multirow{6}{*}{QUARTZ} & \multirow{6}{*}{Quartz\_Content} & TabPFN-opt & 0.003 & 0.000 & -1.39e+07 \\
 &  & TabPFN-Raw & --- & 0.001 & -1.62e+07 \\
 &  & Ridge & 0.000 & \textbf{0.000} & \textbf{2.96e-02} \\
 &  & PLS & \textbf{0.000} & 0.000 & 0.00e+00 \\
 &  & CNN & 0.016 & 0.011 & -3.22e+08 \\
 &  & Catboost & 0.007 & 0.007 & -2.03e+08 \\
\addlinespace[0.8em]
\midrule[\heavyrulewidth]
\multirow{6}{*}{TABLET} & \multirow{6}{*}{Tablet\_ActiveSubstance} & TabPFN-opt & \textbf{0.196} & 0.328 & 2.735 \\
 &  & TabPFN-Raw & --- & 0.843 & -149.764 \\
 &  & Ridge & 0.309 & 0.379 & -12.281 \\
 &  & PLS & 0.309 & 0.338 & 0.000 \\
 &  & CNN & 0.305 & 0.320 & 5.365 \\
 &  & Catboost & 0.245 & \textbf{0.311} & \textbf{7.978} \\
\addlinespace[0.8em]
\midrule[\heavyrulewidth]
\multirow{12}{*}{WOOD\_density} & \multirow{6}{*}{Wood\_Density} & TabPFN-opt & 0.113 & \textbf{0.122} & \textbf{5.237} \\
 &  & TabPFN-Raw & --- & 0.127 & 0.817 \\
 &  & Ridge & 0.112 & 0.128 & 0.393 \\
 &  & PLS & 0.114 & 0.128 & 0.000 \\
 &  & CNN & 0.116 & 0.135 & -5.137 \\
 &  & Catboost & \textbf{0.111} & 0.127 & 1.301 \\
\addlinespace[0.8em]
 & \multirow{6}{*}{Wood\_N} & TabPFN-opt & \textbf{0.046} & \textbf{0.047} & \textbf{6.778} \\
 &  & TabPFN-Raw & --- & 0.048 & 4.517 \\
 &  & Ridge & 0.047 & 0.049 & 3.140 \\
 &  & PLS & 0.047 & 0.051 & 0.000 \\
 &  & CNN & 0.052 & 0.065 & -27.836 \\
 &  & Catboost & 0.046 & 0.051 & -0.358 \\
\addlinespace[0.8em]
\bottomrule
\end{longtable}
\endgroup

\endgroup
\end{landscape}

\clearpage 
\begin{landscape}
\subsection{Complete datasetwise classification results}\label{supp:complete_results_classif}

\begingroup
\setlength{\LTleft}{0pt}
\setlength{\LTright}{0pt}
\begingroup
\small
\setlength{\LTleft}{0pt}
\setlength{\LTright}{0pt}
\setlength{\tabcolsep}{4pt}
\renewcommand{\arraystretch}{1.15}
\begin{longtable}{L{4.8cm}L{8.8cm}L{2.3cm}C{1.7cm}C{1.7cm}C{2.0cm}}
\caption{Dataset-wise classification performance. Only datasets for which the PLS-DA row contains valid results are reported. Best ACC-CV, ACCP, and relative gain versus PLS-DA (highest values) are highlighted in bold.}\label{tab:supp_classif_full}\\
\toprule
Database & Dataset & Model & ACC-CV & ACCP & Rel. ACC gain (\%) \\
\midrule
\endfirsthead
\multicolumn{6}{l}{\textit{Table \thetable{} continued from previous page}}\\
\toprule
Database & Dataset & Model & ACC-CV & ACCP & Rel. ACC gain (\%) \\
\midrule
\endhead
\bottomrule
\endlastfoot
\multirow{10}{*}{ARABIDOPSIS\_CEFE} & \multirow{5}{*}{Arabidopsis\_Genotype} & TabPFN-opt & 0.251 & \textbf{0.396} & \textbf{49.910} \\
 &  & TabPFN-Raw & --- & 0.285 & 7.935 \\
 &  & Catboost & \textbf{0.330} & 0.319 & 20.859 \\
 &  & PLS-DA & 0.303 & 0.264 & 0.000 \\
 &  & CNN & 0.102 & 0.155 & -41.235 \\
\addlinespace[0.8em]
 & \multirow{5}{*}{Arabidopsis\_GrowingCondition} & TabPFN-opt & \textbf{1.000} & \textbf{1.000} & \textbf{0.000} \\
 &  & TabPFN-Raw & --- & \textbf{1.000} & \textbf{0.000} \\
 &  & Catboost & 0.991 & 0.984 & -1.613 \\
 &  & PLS-DA & \textbf{1.000} & \textbf{1.000} & \textbf{0.000} \\
 &  & CNN & 0.500 & 0.500 & -50.000 \\
\midrule[\heavyrulewidth]
\multirow{4}{*}{BEEF\_Impurity} & \multirow{4}{*}{Beef\_Purity} & TabPFN-opt & 0.667 & 0.767 & -14.815 \\
 &  & TabPFN-Raw & --- & \textbf{0.900} & \textbf{0.000} \\
 &  & Catboost & 0.867 & 0.700 & -22.222 \\
 &  & PLS-DA & \textbf{0.933} & \textbf{0.900} & \textbf{0.000} \\
\midrule[\heavyrulewidth]
\multirow{5}{*}{COFFEE\_orig} & \multirow{5}{*}{Coffee\_Origin} & TabPFN-opt & 0.821 & 0.857 & -14.286 \\
 &  & TabPFN-Raw & --- & 0.762 & -23.810 \\
 &  & Catboost & 0.917 & 0.810 & -19.048 \\
 &  & PLS-DA & \textbf{0.960} & \textbf{1.000} & \textbf{0.000} \\
 &  & CNN & 0.143 & 0.143 & -85.714 \\
\midrule[\heavyrulewidth]
\multirow{4}{*}{COFFEE\_sp} & \multirow{4}{*}{Coffee\_Species} & TabPFN-opt & \textbf{1.000} & \textbf{1.000} & \textbf{0.000} \\
 &  & TabPFN-Raw & --- & \textbf{1.000} & \textbf{0.000} \\
 &  & Catboost & \textbf{1.000} & 0.962 & -3.846 \\
 &  & PLS-DA & \textbf{1.000} & \textbf{1.000} & \textbf{0.000} \\
\midrule[\heavyrulewidth]
\multirow{10}{*}{FUSARIUM} & \multirow{5}{*}{Fusarium\_Healthy\_FinalScore} & TabPFN-opt & \textbf{0.854} & 0.702 & -1.883 \\
 &  & TabPFN-Raw & --- & \textbf{0.718} & \textbf{0.448} \\
 &  & Catboost & 0.817 & 0.717 & 0.269 \\
 &  & PLS-DA & 0.849 & 0.715 & 0.000 \\
 &  & CNN & 0.500 & 0.500 & -30.076 \\
\addlinespace[0.8em]
 & \multirow{5}{*}{Fusarium\_Healthy\_Score} & TabPFN-opt & \textbf{0.913} & 0.732 & -3.296 \\
 &  & TabPFN-Raw & --- & 0.662 & -12.564 \\
 &  & Catboost & 0.873 & 0.649 & -14.258 \\
 &  & PLS-DA & 0.906 & \textbf{0.757} & \textbf{0.000} \\
 &  & CNN & 0.500 & 0.500 & -33.951 \\
\midrule[\heavyrulewidth]
\multirow{5}{*}{FruitPuree} & \multirow{5}{*}{FruitPuree\_Strawberry} & TabPFN-opt & \textbf{0.991} & 0.984 & 0.314 \\
 &  & TabPFN-Raw & --- & \textbf{0.986} & \textbf{0.566} \\
 &  & Catboost & 0.976 & 0.983 & 0.252 \\
 &  & PLS-DA & 0.975 & 0.981 & 0.000 \\
 &  & CNN & 0.927 & 0.973 & -0.825 \\
\midrule[\heavyrulewidth]
\multirow{10}{*}{MALARIA} & \multirow{5}{*}{Malaria\_Oocyst} & TabPFN-opt & \textbf{1.000} & \textbf{0.538} & \textbf{0.000} \\
 &  & TabPFN-Raw & --- & 0.509 & -5.263 \\
 &  & Catboost & 0.978 & \textbf{0.538} & \textbf{0.000} \\
 &  & PLS-DA & \textbf{1.000} & \textbf{0.538} & \textbf{0.000} \\
 &  & CNN & 0.500 & 0.500 & -7.018 \\
\addlinespace[0.8em]
 & \multirow{5}{*}{Malaria\_Sporozoite} & TabPFN-opt & \textbf{1.000} & 0.443 & -1.611 \\
 &  & TabPFN-Raw & --- & \textbf{0.595} & \textbf{32.138} \\
 &  & Catboost & \textbf{1.000} & 0.401 & -10.981 \\
 &  & PLS-DA & \textbf{1.000} & 0.450 & 0.000 \\
 &  & CNN & 0.695 & 0.500 & 11.127 \\
\midrule[\heavyrulewidth]
\multirow{4}{*}{MILK} & \multirow{4}{*}{Milk\_Ratio} & TabPFN-opt & \textbf{1.000} & \textbf{1.000} & \textbf{25.000} \\
 &  & TabPFN-Raw & --- & \textbf{1.000} & \textbf{25.000} \\
 &  & Catboost & 0.982 & \textbf{1.000} & \textbf{25.000} \\
 &  & PLS-DA & 0.809 & 0.800 & 0.000 \\
\midrule[\heavyrulewidth]
\multirow{5}{*}{PISTACIA} & \multirow{5}{*}{Pistacia\_Species} & TabPFN-opt & 0.789 & 0.707 & 1.961 \\
 &  & TabPFN-Raw & --- & 0.716 & 3.203 \\
 &  & Catboost & \textbf{0.994} & \textbf{0.737} & \textbf{6.287} \\
 &  & PLS-DA & 0.972 & 0.694 & 0.000 \\
 &  & CNN & 0.668 & 0.200 & -71.173 \\
\end{longtable}
\endgroup

\endgroup
\end{landscape}

\FloatBarrier

\clearpage
\begin{landscape}
\subsection{Complete dataset overview}\label{supp:dataset_overview}

\begingroup
\setlength{\LTleft}{0pt}
\setlength{\LTright}{0pt}
%

\begingroup
\footnotesize
\setlength{\tabcolsep}{3.0pt}
\renewcommand{\arraystretch}{1.15}
\renewcommand{\cellalign}{tl}
\renewcommand{\theadalign}{tl}
\setlength{\LTpre}{4pt}
\setlength{\LTpost}{4pt}
\sloppy
\Urlmuskip=0mu plus 2mu
\urlstyle{same}


\begin{longtable}{L{3.5cm}L{7.9cm}L{2.7cm}L{1.7cm}L{1.95cm}L{1.05cm}R{0.70cm}R{0.70cm}R{0.70cm}R{0.70cm}R{0.85cm}R{0.70cm}}
\caption{\textbf{Regression.} Benchmark overview for regression datasets. Source URLs of the open access databases are given. \textit{p} stands for the number of variables in the dataset, \textit{Outl. test} for Outliers percentage test, \textit{Ext. test} for Extrapolation test. Outlier percentages are based on Hotelling $T^2$ detection on the test set, and extrapolation counts denote the number of test targets outside the support of the train set.}
\label{tab:dataset_description_long_regression} \\
\toprule
\makecell[l]{\bfseries Database\\ } &
\makecell[l]{\bfseries Dataset\\name} &
\makecell[l]{\bfseries Source\\URL} &
\makecell[l]{\bfseries Sample\\type} &
\makecell[l]{\bfseries Target\\trait} &
\makecell[l]{\bfseries Split\\type} &
\makecell[r]{\bfseries N\\total} &
\makecell[r]{\bfseries N\\train} &
\makecell[r]{\bfseries N\\test} &
\makecell[r]{\bfseries p} &
\makecell[r]{\bfseries Outl.\\test} &
\makecell[r]{\bfseries Ext.\\test} \\
\midrule
\endfirsthead

\toprule
\makecell[l]{\bfseries Database\\ } &
\makecell[l]{\bfseries Dataset\\name} &
\makecell[l]{\bfseries Source\\URL} &
\makecell[l]{\bfseries Sample\\type} &
\makecell[l]{\bfseries Target\\trait} &
\makecell[l]{\bfseries Split\\type} &
\makecell[r]{\bfseries N\\total} &
\makecell[r]{\bfseries N\\train} &
\makecell[r]{\bfseries N\\test} &
\makecell[r]{\bfseries p} &
\makecell[r]{\bfseries Outl.\\test} &
\makecell[r]{\bfseries Ext.\\test} \\
\midrule
\endhead

\midrule
\multicolumn{12}{r}{Continued on next page} \\
\midrule
\endfoot

\bottomrule
\endlastfoot

ALPINE
& ALPINE\_P
& \url{https://doi.org/10.18710/CXRCUW}
& Ground dried leaves & P & KS
& 291 & 247 & 44 & 2151 & - & 1 \\
\midrule
AMYLOSE
& Rice\_Amylose
& \url{https://doi.org/10.1016/j.dib.2017.09.077}
& Rice Flour & Amylose content & Y sorted
& 313 & 203 & 110 & 1154 & 3.6\% & 0 \\
\midrule
BEEFMARBLING
& Beef\_Marbling
& \url{https://doi.org/10.57745/FRDOJC}
& Fresh beef carcasse muscle & Marbling & Random
& 832 & 554 & 278 & 331 & 7.9\% & 0 \\
\midrule
\multirow{2}{*}{BEER}
& Beer\_OriginalExtract\_KS
& \url{https://github.com/nanxstats/OHPL/raw/master/data/beer.rda}
& Beer & Original extract & KS
& 60 & 40 & 20 & 576 & - & 0 \\
& Beer\_OriginalExtract\_YBasedSplit
&
& Beer & Original extract & Y sorted
& 60 & 40 & 20 & 576 & - & 0 \\
\midrule
\multirow{3}{*}{BERRY}
& Berry\_Brix
& \url{https://github.com/WongCYS/grapevine_RMI_2025}
& Fresh leaf & Winegrape berry brix & Stratified
& 2133 & 1434 & 699 & 2101 & 10.2\% & 2 \\
& Berry\_pH
&
& Fresh leaf & Winegrape berry pH & Stratified
& 1401 & 912 & 489 & 2101 & 18.4\% & 0 \\
& Berry\_TartaricAcid
&
& Fresh leaf & Winegrape berry tartaric acid content & Stratified
& 1401 & 912 & 489 & 2101 & 18.4\% & 0 \\
\midrule
\multirow{2}{*}{BISCUIT}
& Biscuit\_Fat
& \url{https://rdrr.io/cran/fds/man/Biscuit.html}
& Biscuit dough & Fat content & Random
& 72 & 40 & 32 & 700 & - & 2 \\
& Biscuit\_Sucrose
&
& Biscuit dough & Sucrose content & Random
& 72 & 40 & 32 & 700 & - & 1 \\
\midrule
\multirow{3}{*}{COLZA}
& Colza\_C
& \url{https://doi.org/10.57745/6VYUQN}
& Oilseed rape plant tissues & C content & not specified
& 2419 & 1210 & 1209 & 1154 & 2.6\% & 1 \\
& Colza\_N\_wOutlier
&
& Oilseed rape plant tissues & N content & not specified
& 2427 & 1220 & 1207 & 1154 & 2.9\% & 0 \\
& Colza\_N\_woOutlier
&
& Oilseed rape plant tissues & N content & not specified
& 2412 & 1205 & 1207 & 1154 & 3.1\% & 0 \\
\midrule
\multirow{2}{*}{CORN}
& Corn\_Oil
& \url{https://eigenvector.com/resources/data-sets/}
& Corn kernel & Oil content & Y sorted
& 80 & 64 & 16 & 700 & - & 0 \\
& Corn\_Starch
&
& Corn kernel & Starch content & Y sorted
& 80 & 64 & 16 & 700 & 6.2\% & 0 \\
\midrule
\multirow{3}{*}{DIESEL}
& Diesel\_bp50\_b-a
& \url{https://eigenvector.com/resources/data-sets/}
& Diesel fuel & Boiling point at 50\% recovery & not specified
& 226 & 113 & 113 & 401 & 7.1\% & 0 \\
& Diesel\_bp50\_hla-b
&
& Diesel fuel & Boiling point at 50\% recovery & not specified
& 246 & 133 & 113 & 401 & 8.8\% & 2 \\
& Diesel\_bp50\_hlb-a
&
& Diesel fuel & Boiling point at 50\% recovery & not specified
& 246 & 133 & 113 & 401 & 5.3\% & 0 \\
\midrule
\multirow{4}{*}{DarkResp}
& DarkResp\_SiteCB
& \url{https://doi.org/10.1111/nph.20267}
& Forest tree fresh leaf & Dark respiration & Site
& 470 & 324 & 146 & 2151 & 4.1\% & 1 \\
& DarkResp\_SiteGT
&
& Forest tree fresh leaf & Dark respiration & Site
& 470 & 297 & 173 & 2151 & 86.7\% & 1 \\
& DarkResp\_SiteXSBN
&
& Forest tree fresh leaf & Dark respiration & Site
& 470 & 319 & 151 & 2151 & 11.9\% & 0 \\
& DarkResp\_spxy
&
& Forest tree fresh leaf & Dark respiration & spxy
& 470 & 329 & 141 & 2151 & 1.4\% & 0 \\
\midrule
\multirow{3}{*}{ECOSIS\_LeafTraits}
& EcosisLeaf\_Carotenoid
& \url{https://github.com/UW-GCRL/PLSR_trait_models_evaluation}
& Fresh leaf & Total carotenoid content & spatial
& 4245 & 1016 & 3229 & 196 & 9.0\% & 0 \\
& EcosisLeaf\_Chlorophyll\_SpatialSplit
&
& Fresh leaf & Chla + b & spatial
& 6850 & 2925 & 3925 & 196 & 8.5\% & 0 \\
& EcosisLeaf\_Chlorophyll\_SpeciesSplit
&
& Fresh leaf & Chla + b & species
& 6850 & 3734 & 3116 & 196 & 14.0\% & 0 \\
\midrule
FUSARIUM
& Fusarium\_FvFm
& \url{https://doi.org/10.5281/zenodo.16217833}
& Fresh leaf & Photochemical potential (Fv/Fm) & Group stratified
& 518 & 351 & 167 & 2177 & 10.2\% & 1 \\
\midrule
GRAPEVINES
& Grapevines\_Chloride
& \url{https://github.com/diazgarcialab/grapevine-chloride-prediction}
& Fresh leaf & Leaf chloride content & ks
& 555 & 388 & 167 & 1023 & - & 0 \\
\midrule
\multirow{6}{*}{GRAPEVINE\_LeafTraits}
& GrapevineLeaf\_NetCO2\_ASD
& \url{https://doi.org/10.57745/WVAPOL}
& Dried leaf & Net CO2 assimilation & spxy
& 112 & 78 & 34 & 2101 & 11.8\% & 2 \\
& GrapevineLeaf\_NetCO2\_MicroNIR
&
& Fresh leaf & Net CO2 assimilation & spxy
& 116 & 81 & 35 & 125 & 5.7\% & 3 \\
& GrapevineLeaf\_NetCO2\_MicroNIR\_NeoSpectra
&
& Fresh leaf & Net CO2 assimilation & spxy
& 115 & 80 & 35 & 276 & 5.7\% & 3 \\
& GrapevineLeaf\_NetCO2\_NeoSpectra
&
& Fresh leaf & Net CO2 assimilation & spxy
& 119 & 82 & 37 & 257 & 8.1\% & 4 \\
& GrapevineLeaf\_LMA
&
& Dried leaf & Leaf mass per area & spxy
& 1564 & 1092 & 472 & 2101 & 9.1\% & 2 \\
& GrapevineLeaf\_WUE
&
& Fresh leaf & Water use efficiency & spxy
& 112 & 77 & 35 & 276 & 2.9\% & 0 \\
\midrule
IncombustibleMaterial
& Incombustible\_TIC
& \url{https://github.com/nevernervous78/nirpyresearch/tree/master/data}
& Incombustible material & Total incombustible content & spxy
& 62 & 43 & 19 & 254 & - & 0 \\
\midrule
\multirow{2}{*}{LUCAS}
& Lucas\_SOC
& \url{https://esdac.jrc.ec.europa.eu/content/lucas2015-topsoil-data#tabs-0-description=0}
& Dried topsoil & Soil organic content & ks
& 8731 & 6111 & 2620 & 4200 & 0.8\% & 0 \\
& Lucas\_pH
&
& Dried topsoil & Soil pH & random
& 1763 & 1175 & 588 & 4200 & 5.6\% & 0 \\
\midrule
\multirow{5}{*}{MANURE21}
& Manure\_CaO
& \url{https://doi.org/10.15454/JIGO8R}
& Cattle manure & CaO content & strat spxy
& 490 & 343 & 147 & 1003 & 1.4\% & 0 \\
& Manure\_K2O
&
& Cattle manure & K2O content & strat spxy
& 490 & 343 & 147 & 1003 & 2.0\% & 0 \\
& Manure\_MgO
&
& Cattle manure & MgO content & strat spxy
& 490 & 343 & 147 & 1003 & 3.4\% & 0 \\
& Manure\_P2O5
&
& Cattle manure & P2O5 content & strat spxy
& 490 & 343 & 147 & 1003 & 3.4\% & 0 \\
& Manure\_N
&
& Cattle manure & N content & strat spxy
& 490 & 343 & 147 & 1003 & 4.8\% & 0 \\
\midrule
\multirow{3}{*}{MILK}
& Milk\_Fat
& \url{https://zenodo.org/records/8263430}
& Milk & Fat content & ks
& 402 & 181 & 221 & 255 & 0.5\% & 0 \\
& Milk\_Lactose
& \url{https://zenodo.org/records/8263431}
& Milk & Lactose content & ks
& 1224 & 856 & 368 & 255 & 0.5\% & 0 \\
& Milk\_Urea
& \url{https://zenodo.org/records/8263432}
& Milk & Urea content & ks
& 1224 & 856 & 368 & 255 & 0.5\% & 1 \\
\midrule
\multirow{5}{*}{PHOSPHORUS}
& Phosphorus\_LP
& \url{https://doi.org/10.6084/m9.figshare.28675304}
& Fresh leaf & Lipid P content & spxy by species
& 257 & 169 & 88 & 2101 & 13.6\% & 9 \\
& Phosphorus\_MP
&
& Fresh leaf & Metabolite P content & spxy by species
& 257 & 169 & 88 & 2101 & 13.6\% & 4 \\
& Phosphorus\_NP
&
& Fresh leaf & Nucleic acid P content & spxy by species
& 257 & 169 & 88 & 2101 & 13.6\% & 6 \\
& Phosphorus\_Pi
&
& Fresh leaf & Orthophosphate P content & spxy by species
& 257 & 169 & 88 & 2101 & 13.6\% & 8 \\
& Phosphorus\_V25
&
& Fresh leaf & Photosynthetic capacity (Vcmax25) & spxy by species
& 250 & 168 & 82 & 2101 & 14.6\% & 3 \\
\midrule
PLUMS
& Plums\_Firmness
& \url{https://github.com/nevernervous78/nirpyresearch/blob/master/data}
& Plum & Firmness & spxy
& 40 & 28 & 12 & 600 & - & 0 \\
\midrule
QUARTZ
& Quartz\_Content
& \url{https://github.com/nevernervous78/nirpyresearch/blob/master/data}
& Mineral & Quartz content & spxy
& 303 & 212 & 91 & 1500 & 4.4\% & 0 \\
\midrule
TABLET
& Tablet\_ActiveSubstance
& \url{https://ucphchemometrics.com/tablet/}
& Tablet & Active Substance & ks
& 310 & 207 & 103 & 404 & 44.7\% & 2 \\
\midrule
\multirow{2}{*}{WOOD\_density}
& Wood\_Density
& \url{https://doi.org/10.34725/DVN/24522}
& Ground dried wood auger cores & Wood density & ks
& 402 & 216 & 186 & 1038 & 7.5\% & 3 \\
& Wood\_N
&
& Ground dried wood auger cores & N content & ks
& 402 & 216 & 186 & 1038 & 7.5\% & 1 \\

\end{longtable}


\begin{longtable}{L{3.2cm}L{6.0cm}L{2.7cm}L{1.75cm}L{1.85cm}L{1.05cm}R{0.72cm}R{0.72cm}R{0.72cm}R{0.82cm}R{0.68cm}R{0.82cm}R{0.82cm}}
\caption{\textbf{Classification.} Benchmark overview for classification datasets. Source URLs of the open access databases are given when available. \textit{p} stands for the number of variables in the dataset, \textit{Class imb}. for class imbalance, and \textit{Maj. class} for the proportion of the majority class in the dataset.}
\label{tab:dataset_description_long_classification} \\
\toprule
\makecell[l]{\bfseries Database\\ } &
\makecell[l]{\bfseries Dataset\\name} &
\makecell[l]{\bfseries Source\\URL} &
\makecell[l]{\bfseries Sample\\type} &
\makecell[l]{\bfseries Target\\trait} &
\makecell[l]{\bfseries Split\\type} &
\makecell[r]{\bfseries N\\total} &
\makecell[r]{\bfseries N\\train} &
\makecell[r]{\bfseries N\\test} &
\makecell[r]{\bfseries p} &
\makecell[r]{\bfseries N\\class} &
\makecell[r]{\bfseries Class\\imb.} &
\makecell[r]{\bfseries Maj.\\class} \\
\midrule
\endfirsthead

\toprule
\makecell[l]{\bfseries Database\\ } &
\makecell[l]{\bfseries Dataset\\name} &
\makecell[l]{\bfseries Source\\URL} &
\makecell[l]{\bfseries Sample\\type} &
\makecell[l]{\bfseries Target\\trait} &
\makecell[l]{\bfseries Split\\type} &
\makecell[r]{\bfseries N\\total} &
\makecell[r]{\bfseries N\\train} &
\makecell[r]{\bfseries N\\test} &
\makecell[r]{\bfseries p} &
\makecell[r]{\bfseries N\\class} &
\makecell[r]{\bfseries Class\\imb.} &
\makecell[r]{\bfseries Maj.\\class} \\
\midrule
\endhead

\midrule
\multicolumn{13}{r}{Continued on next page} \\
\midrule
\endfoot

\bottomrule
\endlastfoot

\multirow{2}{*}{ARABIDOPSIS\_CEFE}
& Arabidopsis\_Genotype & \url{https://doi.org/10.1038/s41597-023-02189-w} & Fresh leaf & Genotype group & random blocks & 2185 & 1530 & 655 & 2152 & 10 & 3.26 & 17.6\% \\
& Arabidopsis\_GrowingCondition &  & Fresh leaf & Growing indoor vs outdoor & random blocks & 1263 & 884 & 379 & 2152 & 2 & 1.44 & 59.0\% \\
\midrule
BEEF\_Impurity
& Beef\_Purity & \url{https://www.timeseriesclassification.com/description.php?Dataset=Beef} & Raw and cooked beef & Pure vs adulterated meat & not specified & 60 & 30 & 30 & 470 & 5 & 1.00 & 20.0\% \\
\midrule
COFFEE\_orig
& Coffee\_Origin & \url{https://nirpyresearch.com/analysis-ground-coffee-nir-spectroscopy/} & Aldi Expressi coffee capsules & Coffee origin & strat KS & 70 & 49 & 21 & 601 & 7 & 1.00 & 14.3\% \\
\midrule
COFFEE\_sp
& Coffee\_Species & \url{https://www.timeseriesclassification.com/description.php?Dataset=Coffee} & Freeze-dried coffee & Coffee variety & not specified & 56 & 28 & 28 & 286 & 2 & 1.07 & 51.8\% \\
\midrule
\multirow{2}{*}{FUSARIUM}
& Fusarium\_Healthy\_FinalScore & \url{https://zenodo.org/records/16217833} & Fresh strawberry plant leaf & Healthy or diseased & Stratified & 935 & 646 & 289 & 2177 & 2 & 1.39 & 58.2\% \\
& Fusarium\_Healthy\_Score &  & Fresh strawberry plant leaf & Healthy or diseased & Stratified & 816 & 578 & 238 & 2177 & 2 & 3.27 & 76.6\% \\
\midrule
FruitPuree
& FruitPuree\_Strawberry & \url{https://www.timeseriesclassification.com/description.php?Dataset=Strawberry} & Fruit purée & Strawberry or non-strawberry & ks & 983 & 666 & 317 & 235 & 2 & 1.80 & 64.3\% \\
\midrule
\multirow{2}{*}{MALARIA}
& Malaria\_Oocyst & \url{https://doi.org/10.7910/DVN/YD34OX} & Mosquito & Infected (oocyst) & not specified & 333 & 227 & 106 & 2151 & 2 & 1.09 & 52.3\% \\
& Malaria\_Sporozoite &  & Mosquito & Infected (sporozoite) & not specified & 229 & 138 & 91 & 2151 & 2 & 1.52 & 60.3\% \\
\midrule
MILK
& Milk\_Ratio & \url{https://nirpyresearch.com/detecting-lactose-milk-spectroscopy/} & Milk/lactose-free milk & Ratio milk/lactose-free & strat KS & 450 & 315 & 135 & 601 & 9 & 1.00 & 11.1\% \\
\midrule
PISTACIA
& Pistacia\_Species & \url{https://doi.org/10.18167/DVN1/J1TZZN} & herbarium specimens of pistacia & pistacia species & not specified & 7323 & 5103 & 2220 & 1951 & 5 & 5.26 & 30.9\% \\

\end{longtable}
\endgroup

\endgroup
\end{landscape}

\end{document}